\documentclass[sigconf]{acmart}

\renewcommand\footnotetextcopyrightpermission[1]{}
\definecolor{rblue}{rgb}{0,0.5,1}
\usepackage{times}
\usepackage{graphicx}
\usepackage{amsmath}
\usepackage{booktabs}
\usepackage{multirow}
\usepackage{colortbl}
\usepackage{booktabs}
\usepackage{hhline}
\usepackage{subcaption}

\usepackage{tikz}
\usepackage{lipsum}
\usepackage{paralist}
\usepackage{color}
\newcommand*\circled[1]{\tikz[baseline=(char.base)]{
\node[shape=circle,draw=purple,fill=purple,inner sep=0.6pt] (char) {\textcolor{white}{\footnotesize \textbf{#1}}};}}
\newcommand*\circledmagenta[1]{\tikz[baseline=(char.base)]{
\node[shape=circle,draw=magenta,fill=white,inner sep=0.6pt] (char) {\textcolor{magenta}{\footnotesize \textbf{#1}}};}}
\usetikzlibrary{calc}
\usepackage{longtable}

\newcommand{\repeatcommand}[2]{%
  \ifnum#1>0
    #2%
    \repeatcommand{\numexpr#1-1\relax}{#2}%
  \fi
}

\newcommand{\remdash}[1]{%
  \repeatcommand{#1}{\textemdash}%
}

\newcommand{\ie}{\emph{i.e.}}
\newcommand{\eg}{\emph{e.g.}}

\hypersetup{colorlinks, citecolor=rblue}

\begin{document}

\title{Exploring Few-Shot Adaptation for Activity Recognition on Diverse Domains}

\author{Kunyu Peng$^{1}$, Di Wen$^{1}$, David Schneider$^{1}$, Jiaming Zhang$^{1}$, Kailun Yang$^{2,*}$, M. Saquib Sarfraz$^{1,3}$, Rainer Stiefelhagen$^{1}$, and Alina Roitberg$^{4}$}
\affiliation{%
  \institution{$^{1}$Karlsruhe Institute of Technology, $^{2}$Hunan University, $^{3}$Mercedes-Benz Tech Innovation, $^{4}$University of Stuttgart}
  \authornote{Correspondence (e-mail: {\tt kailun.yang@hnu.edu.cn}).}
  \city{}
  \country{}}
\email{}

\renewcommand{\shorttitle}{RelaMiX}
\renewcommand{\shortauthors}{Peng \textit{et al.}}

\begin{abstract}
Domain adaptation is essential for activity recognition to ensure accurate and robust performance across diverse environments, sensor types, and data sources. Unsupervised domain adaptation methods have been extensively studied, yet, they require large-scale unlabeled data from the target domain. In this work, we focus on \emph{Few-Shot Domain Adaptation for Activity Recognition (FSDA-AR)}, which leverages a very small amount of labeled target videos to achieve effective adaptation. This approach is appealing for applications because it only needs a few or even one labeled example per class in the target domain, ideal for recognizing rare but critical activities. However, the existing FSDA-AR works mostly focus on the domain adaptation on sports videos, where the domain diversity is limited. We propose a new FSDA-AR benchmark using five established datasets considering the adaptation on more diverse and challenging domains. Our results demonstrate that FSDA-AR performs comparably to unsupervised domain adaptation with significantly fewer labeled target domain samples. We further propose a novel approach, RelaMiX, to better leverage the few labeled target domain samples as knowledge guidance. RelaMiX encompasses a temporal relational attention network with relation dropout, alongside a cross-domain information alignment mechanism. Furthermore, it integrates a mechanism for mixing features within a latent space by using the few-shot target domain samples. The proposed RelaMiX solution achieves state-of-the-art performance on all datasets within the FSDA-AR benchmark. To encourage future research of few-shot domain adaptation for activity recognition, our code will be publicly available at \url{https://github.com/KPeng9510/RelaMiX}.
\end{abstract}



\maketitle
\section{Introduction}
Domain shifts, \ie, distribution discrepancies between the source domain data and the target domain data, are inevitable in real-world applications. Most of the existing works in the field of human activity recognition are conducted in single domain~\cite{ma2022motion, wu2021spatiotemporal, shu2022expansion, tong2022semi, chen2023agpn, shen2021fexnet}.
The performance of human activity recognition models is strongly affected by, \eg, changes in sensor types and -placements, different room layouts or transitions from synthetic to real examples~\cite{damen2018scaling, roitberg2021let}.

\begin{figure}[t]
\centering
\includegraphics[width=1\linewidth]{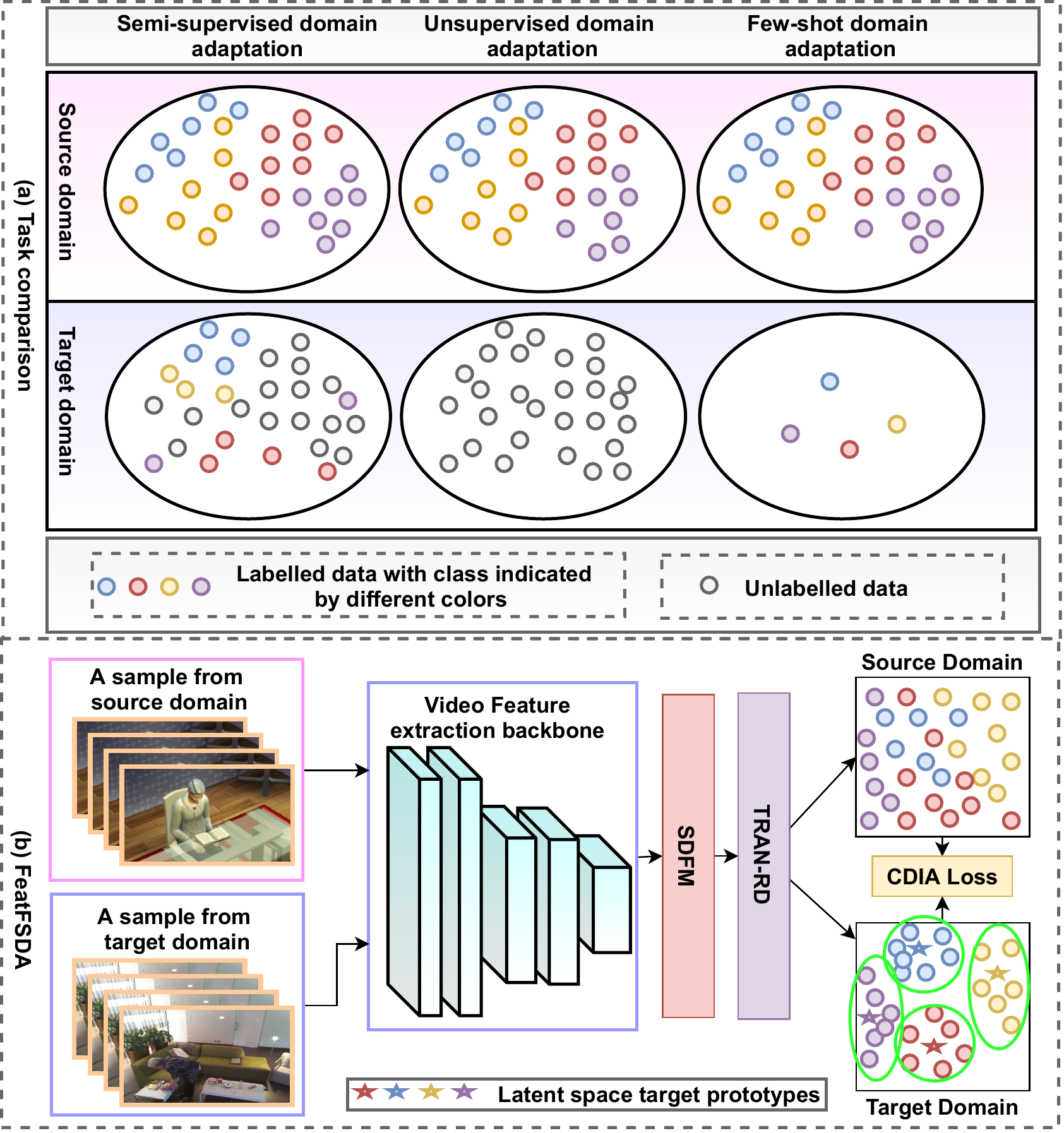}
\vskip-3ex
\caption{(a) Comparison of the Semi-Supervised Domain Adaptation (SSDA), Unsupervised Domain Adaptation (UDA), and Few-Shot Domain Adaptation (FSDA) tasks. (b) An overview of our proposed RelaMiX approach.}
\vskip-4ex
\label{fig:teaser}
\end{figure}

The vast majority of existing domain adaptation research concentrates on the Unsupervised Domain Adaptation (UDA)~\cite{chen2019taaan, wei2022transvae, sahoo2021contrast, ganin2015unsupervised, kang2019contrastive,choi2020shuffle, xu2022learning}, and Semi-Supervised Domain Adaptation (SSDA) settings~\cite{saito2019semi, yang2021deep, yoon2022semi, Li_2021_CVPR} for human activity recognition. 
These paradigms allow transferring the recognition ability from the source domain to the target domain.
The benefit of unsupervised domain adaptation lies in the reduction of labor-intensive labeling tasks for large-scale data in the target domain.
However, as shown in Figure~\ref{fig:teaser}(a), unsupervised domain adaptation and semi-supervised domain adaptation paradigms both require a substantial amount of samples from the target domain, whereas semi-supervised domain adaptation further demands a few labels from the target domain. 

To address this drawback, we take a step towards the area of \textbf{Few-Shot Domain Adaptation for Activity Recognition (FSDA-AR)}, as illustrated in Figure~\ref{fig:teaser}(a). 
Instead of relying on a large number of examples from the target domain, FSDA-AR necessitates a minimal amount of labeled samples, \ie, requiring only a few or even a single annotated sample per class in the target domain. 
These labeled samples from the target domain act as a foundation of knowledge to minimize the domain discrepancy. Contrary to tasks that involve pixel-level detailing, such as semantic segmentation~\cite{fan2020cian}, human activity recognition necessitates only a single label for each video sample. Therefore, annotating a small number of samples in the target domain for FSDA-AR is less labor-intensive. This is because acquiring additional video data requires significant human involvement, either as actors or through video surveillance, making few-shot domain adaptation a more feasible strategy.
Despite this advantage, little research has been performed on FSDA-AR, so far. 

Among the research closely related to FSDA-AR, one study addresses FSDA-AR in videos, however, its benchmark remains unavailable~\cite{gao2020pairwise}. Another investigation~\cite{li2021supervised} is confined to radar-based activity recognition, which is inadequate for comprehensive activity recognition applications. 
Xu~\textit{et al.}~\cite{xu2023augmenting} concentrate on FSDA for sports activities, encountering a lack of domain variability, and their method of feature extraction is not unified as in the UDA task.

In our study, we delve into the versatility of FSDA-AR across a range of complex domain adaptation environments by utilizing an assessment method that includes five publicly accessible video-based human activity recognition datasets. We explore FSDA-AR in varied settings that feature both minor and major domain disparities, assessing transitions from Sims4Action~\cite{roitberg2021let} to ToyotaSmartHome (TSH)\cite{das2019toyota}, EPIC-KITCHEN\cite{damen2018scaling}, HMDB~\cite{kuehne2011hmdb} to UCF~\cite{soomro2012ucf101}, and UCF~\cite{soomro2012ucf101} to HMDB~\cite{kuehne2011hmdb}. This dataset selection spans a range of scenarios from cinematic to real-life third-person views, egocentric activities in kitchens, and synthetic to real domain adaptations, thus maintaining substantial domain variation. Within this benchmark, we include multiple baseline methodologies such as UDA techniques, few-shot activity recognition methods, existing FSDA-AR approaches, and statistical approaches, all adapted for FSDA-AR.

In light of our observations, it has become evident that existing baselines are often incapable of providing consistent performance across the broad spectrum of domains inherent in the FSDA-AR task on most of the challenging domain adaptation datasets.
To address this limitation, there is a pressing need for a novel method tailored specifically to few-shot domain adaptation, one that demonstrates resilience and adaptability to the diverse array of domains encountered in this context. Consequently, we introduce our novel approach for FSDA-AR in this study.
The framework of our approach is guided by three fundamental objectives:
(1) Enhancing temporal data generalization: Our primary goal is to augment the capacity of the model to generalize effectively when dealing with temporal data.
This involves developing mechanisms that facilitate the extraction of temporal patterns and features that are transferrable across different domains.
(2) Leveraging statistical distributions: 
We aim to capitalize on the statistical properties of source-domain samples while effectively utilizing a small number of labeled target-domain samples to enable feature blending within the latent space, thereby enhancing the model's capacity for robust and discriminative feature learning.
(3) Establishing a unified embedding space: We strive to create a shared embedding space that harmoniously integrates both the source and target domains.
This integrated space will enable the model to operate cohesively across the various domains, promoting cross-domain knowledge transfer.

To achieve these objectives, RelaMiX is proposed and its overview is provided in Figure~\ref{fig:teaser}(b).
In our initial endeavor, we introduce a novel Temporal Aggregation Network designed to amplify the generalizability of acquired temporal features. 
This network, termed the Temporal Relational Attention Network with Relation Dropout (TRAN-RD), is purpose-built to capture more nuanced neighborhood information by considering diverse snippet levels, relational attention, and relation combinations, with relation dropout enforced to enhance the representativeness of each relation combination.
Our second innovation centers on a Statistical Distribution-based Feature Mixture (SDFM) mechanism. This mechanism serves to augment the diversity of features within the aligned latent space. By computing the covariance and empirical mean for each temporal snippet, we construct Gaussian distributions for latent space features originating from the source domain.
We subsequently generate mixed-domain features by employing empirical mean transformations and interpolation techniques.
During training, we concurrently fine-tune the temporal aggregation network using features from both the source and target domains, as well as mixed features, fostering cross-domain knowledge transfer.
The Cross-Domain Information Alignment (CDIA) mechanism aligns the source domain with target domain centers and distances them from negative anchors, also applying a similar strategy to mixed features with temporally augmented positives and random mixed negatives. This method bridges the domain gap, enhancing feature transfer using few-shot samples.
These innovations effectively utilize diverse temporal data, enhance feature diversity, and ensure domain alignment, leading to RelaMiX achieving top performance on the benchmark.

Our contributions are summarized as follows:
\begin{compactitem}

  \item We explore the task of Few-Shot Domain Adaptation for Activity Recognition (FSDA-AR) by formalizing a new benchmark considering diverse challenging domains, including movie data to real-world third-person data, cross-person egocentric perspectives, as well as synthetic data to real data. 
  \item We propose the new RelaMiX approach, consisting of three key components: a Temporal Relational Attention Network with Relational Dropout (TRAN-RD) to improve temporal generalizability, a Statistic Distribution-based Feature Mixing (SDFM) to augment the shared latent space, and Cross-Domain Information Alignment (CDIA) to bridge the challenging domain gaps.
  \item Our approach achieves state-of-the art results on the FSDA-AR benchmark considering the $1$-, $5$-, $10$-, $20$-shot settings. Compared with UDA solutions, RelaMiX for FSDA-AR reaches comparable performance. 
\end{compactitem}
\section{Related Work}
In this section, we present an overview of related work in two major areas: domain adaptation and activity recognition.

\noindent\textbf{Domain Adaptation (DA).}
Domain Adaptation (DA)~\cite{chen2019taaan,wei2022transvae,sahoo2021contrast,gao2020pairwise,xiao2021dynamic,li2020enhanced, chang2019domain,xu2023augmenting} refers to a situation in which training and test data come from two domains that are related, but distinct from one another.
Adapting a learner to a target domain is the goal of DA. 
Unsupervised Domain Adaptation (UDA) aims to solve this problem of inter-domain discrepancy without labeling target domain samples in training. Image-based tasks have been widely proposed using UDA methods~\cite{kang2019contrastive, xiao2021dynamic, li2020enhanced, lee2019sliced, long2015learning, long2017deep}. 
For the video-based UDA, several existing works use an adversarial learning framework to deal with the domain shifts~\cite{chen2019taaan,da2022dual}. 
Apart from adversarial learning, Wei~\textit{et al.}~\cite{wei2022transvae} adopt disentanglement learning to decouple the content and context information to achieve a better adaptation. 
CoMix~\cite{sahoo2021contrast} uses mixture techniques to introduce the background information from the target domain into the samples of the source domain during the training procedure to reduce the domain shift.
Besides, recent research works in Semi-Supervised Domain Adaptation (SSDA)~\cite{saito2019semi, yang2021deep, yoon2022semi, Li_2021_CVPR, li2022domain} relax the strict constraint of UDA by using a partially annotated target domain training set.

In Few-Shot Domain Adaptation (FSDA)~\cite{xu2023augmenting, motiian2017few, teshima2020few, jing2023marginalized, huang2021few}, only a few labeled samples per class are given to formulate the target domain training set. 
Note that FSDA-AR does not rely on a large-scale unlabeled training set from the target domain to achieve the domain adaptation, while SSDA needs a large amount of the unlabeled data from the target domain.
Yet, the research of Few-Shot Domain Adaptation in the field of video-based Activity Recognition (FSDA-AR) has been very limited with only three works targeting this task~\cite{gao2020pairwise,li2021supervised, xu2023augmenting}.  
PASTN~\cite{gao2020pairwise} uses an attentive adversarial network to learn domain-invariant features.
FS-ADA~\cite{li2021supervised} integrates both the category classifier and the domain discriminator to extract domain-invariant and category-discriminative features.
Xu~\textit{et al.}~\cite{xu2023augmenting} use the Timesformer~\cite{bertasius2021spacetime} as the backbone and utilize prototype-based snippets contrastive learning to achieve FSDA-AR.

However, the benchmark of PASTN~\cite{gao2020pairwise} is not available, the datasets used by Xu~\textit{et al.}~\cite{xu2023augmenting} do not encompass different levels of the domain shift, and the feature extraction backbone is not unified as the same backbone, \eg, I3D~\cite{carreira2017quo}, which is widely used in UDA.
Besides, FS-ADA~\cite{li2021supervised} is developed for radar data. 
In this regard, we consider that a fair comparison should be conducted with UDA approaches reformulated in the FSDA-AR task for the adaptation against diverse domain shifts. 
Moreover, there is a strong need for further research of video-based FSDA-AR on diverse domain shifts, \eg, different views of egocentric videos and synthesized videos to real videos, taking into account the unification of the feature extraction backbone to make fair task comparison with UDA and the existing works in the FSDA-AR field. 
To address this issue, we introduce a novel video-based FSDA-AR benchmark with diverse domain combinations and under the unification of the feature extraction backbone. 
The RelaMiX approach enhances video data generalization through three components: a Temporal Relational Attention Network with Relation Dropout for better temporal embeddings, a Statistic Distribution-Based Feature Mixture inspired by~\cite{yang2021free} for diversified latent space targeting the domain, and Cross-Domain Information Alignment using contrastive supervision with mixed domain negatives and prototype positives. This method demonstrates strong performance in Few-Shot Domain Adaptation for Activity Recognition (FSDA-AR) across five datasets.

\noindent\textbf{Video-based Human Activity Recognition (HAR).} Supervised human activity recognition methods~\cite{simonyan2014two, wang2016temporal, tran2015learning, carreira2017quo, xie2018rethinking, feichtenhofer2020x3d, peng2022transdarc} have achieved impressive results with deep learning algorithms in recent years. The video-based approaches can be grouped into Convolutional Neural Networks (CNNs) based and transformer-based methods.
For CNN-based methods, most of the existing approaches leverage 3D CNNs and diverse temporal aggregation techniques. 
TSN~\cite{wang2016temporal} samples a fixed number of video frames evenly across the video segments, and uses these sampled frames as input for a two-stream network.
I3D~\cite{carreira2017quo} uses an inflated Inception v1 model~\cite{DBLP:journals/corr/SzegedyLJSRAEVR14} with 3D convolutional layers utilized in each stage. 
Another group of network architectures leverages transformer-based backbones~\cite{fan2021multiscale, bertasius2021spacetime, liu2021video}. Most of existing UDA works~\cite{chen2019taaan, wei2022transvae, sahoo2021contrast, da2022dual} make use of I3D~\cite{carreira2017quo} or ResNet~\cite{DBLP:journals/corr/HeZRS15} as the feature extractor. 
Following the standard setting of UDA~\cite{chen2019taaan, wei2022transvae, sahoo2021contrast, da2022dual}, we leverage I3D~\cite{carreira2017quo} as our feature extractor in the proposed RelaMiX method, allowing adequate comparison between the FSDA-AR and UDA tasks. To enable domain generalizable feature learning, our RelaMiX incorporates generalizable temporal-relation-based aggregation, source domain statistics-based feature mixture, and cross-domain information alignment supervision, which enable RelaMiX to elevate state-of-the-art performances on the FSDA-AR benchmark.

\begin{figure*}[t!]
\centering
\includegraphics[width=1\linewidth]{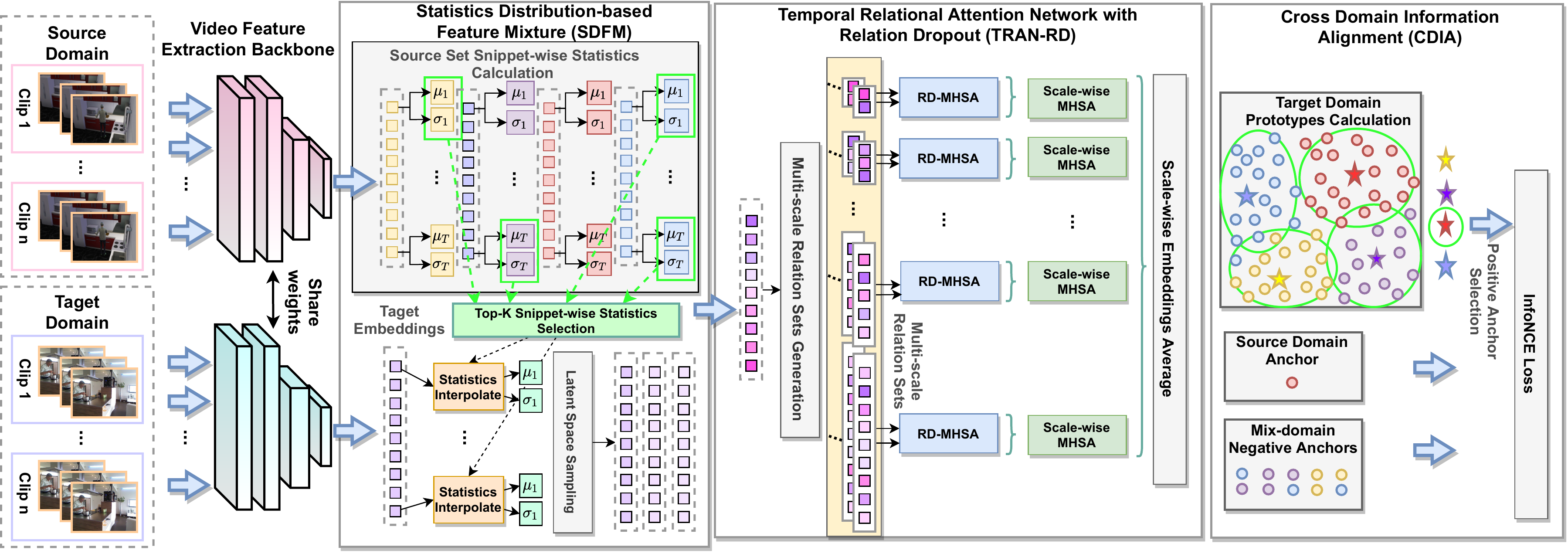}
\vskip-3ex
\caption{The RelaMiX framework processes video by dividing it into overlapping snippets to extract features. It calculates the statistics for these snippets from the source domain, synthesizing cluster centers for the target-domain latent space. Temporal relation sets are refined using Relation-Dropout Multi-Head Self-Attention and Scale-wise Multi-Head Self-Attention for feature learning and aggregation. Cross-Domain Information Alignment (CDIA) loss is used alongside cross-entropy losses to minimize the domain gap.}
\vskip-3ex

\label{fig:main}
\end{figure*}
\section{Method}
\subsection{Problem Formulation}
\label{sec:pf}

The FSDA-AR task assumes a fully labeled large-scale source domain training set $D_{source} = {(v_i^{source}, l_i^{source})}_{i=1}^{N_s^{train}}$ and a small target domain training set $D_{target} = {(v_i^{target}, l_i^{target})}_{i=1}^{N_{target}^{train}}$, which contains only few labeled samples per class. Our objective is to train a model using both $D_{source}$ and $D_{target}$ so that the resulting model performs well on the target domain test set, denoted as $T_{target} = {(v_i^{target}, l_i^{target})}_{i=1}^{N_{target}^{test}}$.
In this context, $N_{source}^{train}$, $N_{target}^{train}$, and $N_{target}^{test}$ represent the number of samples in the source domain training set, target domain training set, and target domain test set, respectively. The index of the sample is indicated by $i$, while the video samples and labels are represented by $v$ and $l$, respectively.

\subsection{Baselines on FSDA-AR Benchmark}
\label{sec:ib}
We construct diverse baselines for our FSDA-AR benchmark by using three FSDA-AR approaches, \ie, 
FS-ADA~\cite{li2021supervised}, SSA$^2$Lign~\cite{xu2023augmenting}, and PASTN~\cite{gao2020pairwise}, 
established UDA approaches reformulated for FSDA-AR, \ie, TA$^3$N~\cite{chen2019taaan} and TranSVAE~\cite{wei2022transvae}, CoMix~\cite{sahoo2021contrast}, and CO$^2$A~\cite{da2022dual}, few-shot human action recognition approaches reformulated for FSDA-AR, \ie, TRX~\cite{perrett2021temporal} and HyRSM~\cite{wang2022hybrid}, as well as the statistical baselines, \ie, random chance (Random), K-Nearest Neighbors (KNN), Nearest Neighbor (NN), and Nearest Center (NC). 
For consistency, we use the I3D backbone~\cite{carreira2017quo} pre-trained on Kinetics400~\cite{kay2017kinetics} for all baselines, as I3D is commonly used for video feature extraction in UDA tasks~\cite{wei2022transvae}.

\noindent\textbf{Statistical Baselines.} 
Video features are extracted with I3D, omitting the final classification layer, and these features underpin all statistical baselines except for the Random method. The Random baseline randomly assigns a class to test samples, setting a performance lower bound.
In the KNN method, labels for test samples are determined by averaging outcomes from the $3$-, $5$-, and $10$-NN methods, using neighbors from the source domain. The Nearest Center method assigns labels based on the nearest class center in the source domain, while the Nearest Neighbor method uses the closest source domain sample for labeling. These methods maintain consistent performance across different shot settings by using the same source domain. They serve as statistical baselines to highlight Domain Generalization (DG) performance, emphasizing the advantage of incorporating few-shot samples from the target domain.

\noindent\textbf{Unsupervised Domain Adaptation (UDA) Baselines.} To enrich our FSDA-AR benchmark and enable comparisons with the existing domain adaptation frameworks, we implement and restructure several established video-based UDA methods to suit the few-shot domain adaptation task.
This is achieved by incorporating supervised classification loss on the target domain training set and converting the unsupervised contrastive loss into triplet margin loss where only few-shot samples from the target domain are provided.
Specifically, we leverage four prominent methods: TA$^3$N~\cite{chen2019taaan}, TranSVAE~\cite{wei2022transvae}, CO$^2$A~\cite{da2022dual}, and CoMix~\cite{sahoo2021contrast}, where TA$^3$N and CO$^2$A leverage domain adversarial learning, TranSVAE concentrates on domain disentanglement, and CoMix targets at bridging the domain gaps by mixing the background information from both domains.

\noindent\textbf{Few-shot Learning Baselines.}
Two representative works for video-based few-shot learning are leveraged, \ie, TRX~\cite{perrett2021temporal} and HyRSM~\cite{wang2022hybrid}, to test the performance of few-shot learning approaches in the few-shot domain adaptation task.

\noindent\textbf{Few-shot Domain Adaptation (FSDA) Baselines.}
We employ three existing approaches of FSDA-AR. 
The first approach is FS-ADA~\cite{li2021supervised}, which targets radar-based FSDA-AR using adversarial domain adaptation. 
The second approach is PASTN~\cite{gao2020pairwise}, which introduces a pairwise attentive, adversarial spatiotemporal network, and the third approach is SSA$^2$Lign~\cite{xu2023augmenting}, which makes use of attentive alignment of snippets to bridge the domain gap.
To explore FSDA-AR against more diverse domain shifts and achieve the unification of the feature backbone to keep a fair comparison with the UDA task, we establish a novel benchmark, using publicly accessible datasets encompassing diverse domain styles, and then replicate their performance by unifying the video extraction backbone as I3D~\cite{carreira2017quo} to provide a fair comparison with UDA methods.

\subsection{Introduction of RelaMiX Method}
\label{sec:RelaMiX}
In this section, we present the key ideas of our proposed RelaMiX approach (depicted in Figure~\ref{fig:main}). 
RelaMiX first uses I3D~\cite{carreira2017quo} to extract snippet-wise features. Then it incorporates a statistic distribution-based feature mixture approach to enrich the information in the latent space shared across domains, a temporal relation attention network with relation dropout to achieve more generalizable temporal information aggregation, and a cross-domain information alignment loss for representative feature learning to bridge the domain gap.

\noindent\textbf{Snippet-wise Video Feature Extraction.}
Similarly to our baselines, we adopt I3D~\cite{carreira2017quo} as our backbone to obtain video representations and a temporal sliding window to extract the snippet features for a given video. Let us consider a video sample represented as $\mathbf{v} = \{ \mathbf{v}_{1}, ..., \mathbf{v}_{N_T}\}$, where $\mathbf{v}_{i}$ indicates the $i$-th frame of the given video $\mathbf{v}$, $N_w$ is the window size of the sliding temporal window and $N_T$ is the number of the frames. We can obtain a set of video snippets, denoted as $\{\mathbf{s}_i\} =\{\{\mathbf{v}_{i}, ..., \mathbf{v}_{i + N_w} \}~|~ i\in \left[ 1, N_T - N_w\right]$\} (zero padding is used at the start and end of the video). Next, we extract snippet-wise features. By inputting these snippets into the previously mentioned I3D-based feature extractor $\mathbf{H}_{\alpha}$, we can acquire the set of clip features for a sample as $\mathbf{f} = \left[\mathbf{H}_\alpha(\mathbf{s}_1),..., \mathbf{H}_\alpha(\mathbf{s}_{N_T - N_w})\right]$.

\noindent\textbf{Statistic Distribution-based Feature Mixture (SDFM).}
To better leverage the information provided by the few-shot samples from the target domain, we propose a new method, SDFM, to synthesize more target-domain embeddings by using the statistics calculated on the snippet features from the source domain training set. Next, the details of the SDFM will be illustrated. We first calculate the statistic centers and covariance matrix as Eq.~\ref{eq:1} and Eq.~\ref{eq:1_1}.
\begin{equation}
\label{eq:1}
    \mathbf{\mu}_{c}^{(t, source)} = \frac{\sum_{i=1}^{N_{c}}{\mathbf{f}_{(i,c)}^{(t, source)}}}{N_c},
\end{equation}
\begin{equation}
\label{eq:1_1}
     \mathbf{\sigma}_{c}^{(t, source)}  = \sqrt{\frac{\sum_{i=1}^{N_{c}}{(\mathbf{f}_{(i,c)}^{(t,source)} - \mathbf{\mu}_{c}^{(t,source)})^2}}{N_c-1}},
\end{equation}
where $\mu_{c}^{(t,source)}$ and $\sigma_{c}^{(t,source)}$ indicate the mean and covariance of embeddings of the $c$-th category and $t$-th snippet from the source domain. 
$N_c$ denotes the sample number for category $c$.
$\mathbf{f}_{(i,c)}^{(t,source)}$ indicates the embeddings from the $i$-th sample from $c$-th category of the $t$-th snippet. 

Then, for a sample provided by the few-shot target domain training set with embeddings $\hat{\mathbf{f}}^{(t,target)}$ of the $t$-th snippet, we first calculate the Top-K nearest cluster centers for each snippet according to its distance to the mean of the snippet embeddings from the source domain individually, as depicted in Eq.~\ref{eq:2}.
\begin{equation}
\label{eq:2}
    I_c^t = TopK_{c \in \Omega_c}(e^{(1-D(\mu_c^{(t, source)} \cdot \mathbf{\hat{f}}^{(t,target)}))}),
\end{equation}
where $I_c^t$ indicates the categories which are selected. $D(\cdot)$ indicates the euclidean distance. $\Omega_c$ indicates the set of categories.
Then we calculate the synthesized empirical mean for this target domain embeddings according to Eq.~\ref{eq:3},
\begin{equation}
\label{eq:3}
    \hat{\mathbf{\mu}}^t, \mathbf{\hat{\sigma}}^t = \frac{\mathbf{\hat{f}}^{(t,target)} + \sum_{k\in I_c^t}{\mu_{k}^{(t, source)}}}{K+1}, \frac{\sum_{k\in I_c^t}{\sigma_{k}^{(t,source)}}}{K} + \alpha,
\end{equation}
where $\alpha$ is a fixed factor and $K$ is the number of selected centers.
Next, we build up a multi-variant normal distribution based on the synthesized empirical mean and covariance according to Eq.~\ref{eq:4}.
\begin{equation}
\label{eq:4}
    \mathbf{\hat{f}}^{(t,target)}_{new} \sim \frac{1}{\hat{\sigma}^t \sqrt{2\pi}} e^{-\frac{( \mathbf{x}- \hat{\mu}^t)^2}{2(\hat{\sigma}^t)^2}}.
\end{equation}
Additional embeddings can be derived by leveraging the established normal distributions for each temporal snippet within the target domain.
The new generated features $\hat{\mathbf{f}}_{new}^{(t,target)}$ share the same class with $\hat{\mathbf{f}}^{(t,target)}$.
This process utilizes statistical parameters obtained from the source domain, in conjunction with the provided few-shot samples from the target domain.
The purpose of this approach is to enhance diversity within the latent space while simultaneously reasoning information from both the source and target domains.

\noindent\textbf{Temporal Relational Attention Network with Relation Dropout (TRAN-RD).} When considering adapting the model into another domain by a few samples, a temporal aggregation approach with great generalizability will help to grasp the important cues from different temporal relations. To delve into this concern, we propose a new temporal aggregation mechanism, \ie, temporal relational attention network with relation dropout, which incorporates two major concepts, \ie, Relation-Dropout based Multi-Head Self-Attention (RD-MHSA) and Scale-wise Multi-Head Self-Attention (Scale wise-MHSA), to attend on different temporal relational granularities, and learn representative and generalizable features. 
RD-MHSA is conducted on all of the embeddings from the source domain, the target domain, and the generated embedding set of the target domain. We first generate multi-scale relational index sets as Eq.~\ref{eq:5}, 
\begin{equation}
\label{eq:5}
\begin{aligned}
    \Omega_r = &\{ (i,...,k,...,j) | 
    i\leq ... \leq k \leq... \leq j,
    \\&(i,..., k,...,j) \in \left[1, N_T\right]^r ,~and~ r\in\left[2, N_T\right] \},
\end{aligned}
\end{equation}
where we can grasp relational indexes according to different scales. The selected snippets preserve the temporal order. $r$ is used to define the selected scales of the desired relational set. 

\noindent\underline{Relation-Dropout Multi-Head Self-Attention (RD-MHSA)} is used first to aggregate the temporal features within each snippet.
The relational attended snippet embedding $\mathbf{\hat{f}}_s$ for one snippet $\mathbf{f}_s$ from the given snippet relation set $\Omega_s \subseteq \Omega_r$ can be calculated through Eq.~\ref{eq:6},
\begin{equation}
\label{eq:6}
    \mathbf{f}_{a} = LN\left[SM\left[\frac{\mathbf{P}_Q(\mathbf{f}_s) \cdot \mathbf{P}_K(\mathbf{f}_s)}{\sqrt{d_k}}\right]*\mathbf{P}_V(\mathbf{f}_s)\right],
\end{equation}
\begin{equation}
    \hat{\mathbf{f}}_{s} = LN\left[\mathbf{f}_s + \sum_{h=1}^{N_h}{\mathbf{f}_a^h} + FFN(\mathbf{f}_s)\right],
\end{equation}
where $LN$ indicates layer normalization, $SM$ indicates the SoftMax operation, and $FFN$ indicates a multi-layer-perception (MLP) based Feed-Forward Network. 
$d_k$ is a scale factor. $\mathbf{P}_Q$, $\mathbf{P}_K$, and $\mathbf{P}_V$ are chosen as linear projections. 
$\mathbf{f}_a^h$ indicates the relational attention obtained through $N_h$ heads. 
After the aforementioned process, we obtain the self-attended relation set $\hat{\Omega}_{s}$.
Then, we apply dropout to the snippets within each attended relation set as Eq.~\ref{eq:8}.
\begin{equation}
\label{eq:8}
    \hat{\Omega}_{s}^{DP}= DropOut(\hat{\Omega}_s, \beta),
\end{equation}
where $\beta$ is the pre-defined dropout ratio for the snippets inside one relation set. The RD-MSHA aims at learning representative features when snippets are randomly not available in the training procedure.

\noindent\underline{Scale-wise Multi-Head Self-Attention (Scale-wise MHSA)} is then used to aggregate the information within each scale. We first do concatenation considering all snippets inside one relation set after the relation dropout along the temporal dimension, denoted as $\hat{\mathbf{f}}_s^{\Omega_s}$. Then we regard the temporal dimension as the token dimension for MHSA. Scale-wise MHSA is then achieved by Eq.~\ref{eq:9} and Eq.~\ref{eq:10}, 
\begin{equation}
\label{eq:9}
    \hat{\mathbf{f}}_{a}^{\Omega_s} = LN\left[SM\left[\frac{\hat{\mathbf{P}}_Q(\hat{\mathbf{f}}_s^{\Omega_s}) \cdot \hat{\mathbf{P}}_K(\hat{\mathbf{f}}_s^{\Omega_s})}{\sqrt{\hat{d}_k}}\right]*\hat{\mathbf{P}}_V(\hat{\mathbf{f}}_s^{\Omega_s})\right],
\end{equation}
\begin{equation}
\label{eq:10}
\widetilde{\mathbf{f}}_{a}^{\Omega_s} = LN\left[\hat{\mathbf{f}}_{s}^{\Omega_s} + \sum_{h=1}^{N_h}{\hat{\mathbf{f}}_{a}^{(\Omega_s,h)}} + FFN(\hat{\mathbf{f}}_{s}^{\Omega_s})\right],
\end{equation}
where all the projections $\hat{\mathbf{P}}_Q$, $\hat{\mathbf{P}}_K$, and $\hat{\mathbf{P}}_V$ are chosen as linear projections. $\hat{d}_k$ is fixed scale factor. $N_h$ denotes the number of the heads. Then we calculate the final aggregated embedding through Eq.~\ref{eq:11}.
\begin{equation}
\label{eq:11}
    \mathbf{f}^{*} = \frac{\sum_{\Omega_s \subseteq \Omega_r}{\widetilde{\mathbf{f}}_{a}^{\Omega_s}}}{N_s},
\end{equation}
where $N_s = N_T-1$ denotes the total number of the relation sets.

\noindent\textbf{Cross Domain Information Alignment (CDIA).} After the enrichment of the latent space features and the development of a generalizable temporal relation aggregation method, we further try to introduce more constraints during learning to bridge the gap between the source domain and target domain by using the provided few-shot target domain samples. 
When a source domain anchor is given as $\mathbf{f}^{source}$ after the TRAN-RD, we wish it could be closer to its corresponding cluster centers $\mathbf{f}^{target}_c$ calculated on the target domain while being far away from the negative anchors $\mathbf{\widetilde{f}}^{source}$ from different categories.
The CDIA loss can be therefore calculated via Eq.~\ref{eq:12},
\begin{equation}
\label{eq:12}
    \mathcal{L}_{CDIA} = -\frac{1}{N} \sum_{i=1}^{N} \log \left[\frac{e^{(cos(\mathbf{f}^{source}_{i},~ \mathbf{f}^{target}_{(i,c)}))}}{\sum_{j=1}^{N_n} e^{(cos(\mathbf{f}^{source}_{i},~\mathbf{\widetilde{f}}^{source}_{j}))}}\right],
\end{equation}
where $N$ denotes the sample number from the source domain and $N_n$ denotes the number of the negative anchors for the $i-$th anchor from the source domain. $cos$ indicates the cosine similarity. $\mathbf{f}^{target}_{(i,c)}$ indicates the nearest target domain center ($c$-th category) for $i-$th anchor. The target domain centers can be calculated by Eq.~\ref{eq:13}.
\begin{equation}
\label{eq:13}
    \mathbf{f}^{target}_{c} = \frac{\sum_{i=1}^{N_c}{ \mathbf{f}^{target}_{(i,c)}}}{N_c},
\end{equation}
where $N_c$ indicates the sample number for the class $c$ and $\mathbf{f}^{target}_{c}$ indicates the target domain center for class $c$.
Apart from the CDIA loss, we make use of supervised cross-entropy losses for both the samples from the source domain training set, target domain few-shot training set, and target domain generated training set, \ie, $L_{CES}$, $L_{CET}$, and $L_{CEA}$. 
To get representative features from the generated target domain training set, we make use of another contrastive learning loss through Eq.~\ref{eq:14}, 
\begin{equation}
\label{eq:14}
    \mathcal{L}_{aux} = -\frac{1}{N_g} \sum_{i=1}^{N_g} \log \left[\frac{e^{(cos(\mathbf{f}^{target}_{i},~ \mathbf{\hat{f}}^{target}_{i}))}}{\sum_{k=1}^{N_n}e^{(cos(\mathbf{f}^{target}_{i},~ \widetilde{\mathbf{f}}^{target}_{k}))}}\right],
\end{equation}
where the positive anchors $\mathbf{\hat{f}}^{target}_i$ are generated through random permutation of the input snippet along the temporal axis while the negative anchor $\widetilde{\mathbf{f}}^{target}_k$ is the sample from different classes. 
$N_g$ indicates the number of the generated features in the shared latent space. 
$cos$ denotes cosine similarity.
The supervision is achieved by a weighted sum of the aforementioned loss functions by Eq.~\ref{eq:15}.

\begin{equation}
\label{eq:15}
\begin{split}
    L_{all} = \omega_1*L_{CDIA} + \omega_2*L_{CES} + \omega_3*L_{CET} + \\ 
    \omega_4*L_{CEA} + \omega_5*\mathcal{L}_{aux}.
\end{split}
\end{equation}
\section{Experiments}
\subsection{Datasets}
\begin{table*}[t]
\caption{%
Experimental results on UCF~\cite{soomro2012ucf101} $\rightarrow$ HMDB~\cite{kuehne2011hmdb},  HMDB~\cite{kuehne2011hmdb} $\rightarrow$ UCF~\cite{soomro2012ucf101}, and EPIC-KITCHEN~\cite{damen2018scaling}. S-X indicates Shot-X.
}
\vskip-2ex
\label{tab:hmdb}
\centering
\newcommand{\emdashnum}{6}
\scalebox{0.74}{\begin{tabular}{ll|llll|llll|llll|llll} 
\toprule
\midrule
& \multirow{2}{*}{\textbf{Method}} & \multicolumn{4}{c|}{\textbf{\circled{1} UCF $\rightarrow$ HMDB}} & \multicolumn{4}{c|}{\textbf{\circled{2} HMDB $\rightarrow$ UCF}} & \multicolumn{4}{c|}{\textbf{\circled{3} EPIC-KITCHEN mean}}& \multicolumn{4}{c}{\textbf{\circled{4} Sims4Action $\rightarrow$ ToyotaSmartHome}} \\ 
&  & \textbf{S-1} & \textbf{S-5} & \textbf{S-10} & \textbf{S-20} &  \textbf{S-1} & \textbf{S-5} & \textbf{S-10} & \textbf{S-20} &  \textbf{S-1} & \textbf{S-5} & \textbf{S-10} & \textbf{S-20}& \textbf{S-1} & \textbf{S-5} & \textbf{S-10} & \textbf{S-20} \\

\midrule
\multirow{4}{*}{\rotatebox[origin=c]{90}{\textsc{DG} }} & Random & \multicolumn{4}{c}{\remdash{\emdashnum}\ \ 8.6 \remdash{\emdashnum}} & \multicolumn{4}{c}{\remdash{\emdashnum}\ \ 8.1 \remdash{\emdashnum}} & \multicolumn{4}{c}{\remdash{\emdashnum}\ 12.5 \remdash{\emdashnum}} & \multicolumn{4}{c}{\remdash{\emdashnum}\ 11.1 
 \remdash{\emdashnum}}\\ 
& KNN & \multicolumn{4}{c}{\remdash{\emdashnum}\ 81.1 \remdash{\emdashnum}}  & \multicolumn{4}{c}{\remdash{\emdashnum}\ 88.3 \remdash{\emdashnum}} & \multicolumn{4}{c}{\remdash{\emdashnum}\ 28.1
 \remdash{\emdashnum}} & \multicolumn{4}{c}{\remdash{\emdashnum}\ \ 3.3 
 \remdash{\emdashnum}}  \\ 
& Nearest Center & \multicolumn{4}{c}{\remdash{\emdashnum}\ 83.9 \remdash{\emdashnum}} & \multicolumn{4}{c}{\remdash{\emdashnum}\ 91.5 \remdash{\emdashnum}} & \multicolumn{4}{c}{\remdash{\emdashnum}\  27.4
 \remdash{\emdashnum}}& \multicolumn{4}{c}{\remdash{\emdashnum}\ 28.0 
 \remdash{\emdashnum}}  \\ 
 & Nearest Neighbor & \multicolumn{4}{c}{\remdash{\emdashnum}\ 80.1 \remdash{\emdashnum}} & \multicolumn{4}{c}{\remdash{\emdashnum}\ 88.6 \remdash{\emdashnum}} & \multicolumn{4}{c}{\remdash{\emdashnum}\  25.5
 \remdash{\emdashnum}}& \multicolumn{4}{c}{\remdash{\emdashnum}\ 3.27
 \remdash{\emdashnum}}  \\ 
\midrule
\multirow{10}{*}{\rotatebox[origin=c]{90}{\textsc{FSDA-AR}}} & CoMix~\cite{sahoo2021contrast} & 83.1 & 88.1 & 89.7 & 90.8  &91.0 & 93.2 & 96.8 &  96.3 & 31.2 & 31.8 & 32.1 &  32.7& 24.2 & 20.6 & 28.9 & 35.5 \\
& CO$^2$A~\cite{da2022dual} & 83.9 & 88.1 & 89.1 & 91.1 & 92.5 & 94.0 & 96.7 &  97.5 & 32.6 & 36.4 & 38.0 & 38.2& 21.9 & 26.6 & 34.0 &  42.1\\
& TA$^3$N~\cite{chen2019taaan} & 83.3 & 88.9 & 88.3 & 91.7& 93.7 & 95.1 & 97.5 & 98.0  & 37.9 & 41.2 & 42.1 & 43.0 & 21.1 & 29.8 & 35.8 &  42.7\\
& TranSVAE~\cite{wei2022transvae}& 82.3 & 82.8 & 83.2 & 84.8 & 89.7 & 89.0 & 94.4 & 95.1  & 37.6 & 41.1 & 40.8 & 43.3 & 22.6 & 22.7 & 18.9 & 22.7 \\
\cmidrule{2-18}
& TRX~\cite{perrett2021temporal} & 77.2 & 80.3 & 78.6 & 81.9& 82.2 & 83.1 & 81.1 &  84.4 & 26.7 & 27.4 & 28.7 & 30.2 & 14.0 & 13.8 & 19.0& 18.9\\
& HyRSM~\cite{wang2022hybrid} & 79.7 & 81.1 & 82.2 & 83.6 & 88.1 & 90.1 & 91.0 & 90.8  & 35.8 & 36.7 & 37.1 & 37.8 & 18.9& 22.4 & 27.4 & 28.0\\
\cmidrule{2-18}
& FS-ADA~\cite{li2021supervised} & 82.7 & 87.2 & 88.6& 87.2 & 91.9& 94.4& 93.7 & 96.5 & 37.0 & 39.7 & 39.3 & 40.4 & 17.1 & 22.6 & 28.3 & 28.0\\
& PASTN~\cite{gao2020pairwise} & 83.4 & 86.2 & 88.3 & 89.8 & 91.2& 94.2& 95.8& 96.5  & 36.1 & 40.5 & 40.3 & 42.5& 22.6 & 22.6 & 22.6& 28.0\\
& SSA$^2$lign~\cite{xu2023augmenting}& 80.6& 85.0 & 88.3 & 87.8 & 87.0 & 94.4 & 94.6 &  94.4 & 31.5 & 40.1 & 40.9 & 42.0 & 22.6 & 23.7 & 35.0& 41.3 \\
\cmidrule{2-18}
\rowcolor{gray!15} & RelaMiX (ours) & \textbf{85.6} & \textbf{91.1} & \textbf{91.1} & \textbf{92.2} & \textbf{94.1} & \textbf{97.2} & \textbf{97.9} &  \textbf{98.4} & \textbf{40.7}& \textbf{43.9} & \textbf{44.4} & \textbf{45.2} & \textbf{27.0} & \textbf{31.0} & \textbf{38.9} & \textbf{49.2}\\
\hline
\bottomrule
\end{tabular}}
\vskip-1ex
\end{table*}
We utilize five popular human activity recognition datasets, \eg, HMDB-51~\cite{kuehne2011hmdb}, UCF-101~\cite{soomro2012ucf101}, EPIC-KITCHENS-55~\cite{damen2018scaling}, ToyotaSmartHome (TSH)~\cite{das2019toyota}, and Sims4Action~\cite{roitberg2021let}, to investigate FSDA-AR. 
The selected datasets comprise diverse activities and recording environments, enabling comprehensive evaluation of DA techniques. 
The group of the methods that are under \textit{DG} only use the samples from the source domain, while the methods that are under \textit{FSDA-AR} make use of the samples from both the source and target domain.

\noindent\textbf{HMDB-51}~\cite{kuehne2011hmdb} contains $6,766$ video clips from various sources, spanning across $51$ activities categories with a minimum of $101$ clips per activity. $12$ action classes are chosen in the DA task.

\noindent\textbf{UCF-101}~\cite{soomro2012ucf101} consists of $13,320$ video clips, divided into $101$ activity categories. These datasets are employed in the HMDB $\rightarrow$ UCF and UCF $\rightarrow$ HMDB adaptation tasks where they share $12$ classes.

\noindent\textbf{EPIC-KITCHENS-55}~\cite{damen2018scaling} comprises $55$ hours of egocentric videos, capturing kitchen activities from $32$ participants, and is utilized in the Epic-Kitchens domain adaptation task. We leverage the domain adaptation benchmark defined by~\cite{munro2020multi} on $8$ overlapped activities.

\noindent\textbf{ToyotSmartHome} (TSH)~\cite{das2019toyota} includes $16,115$ video clips, containing $31$ daily living activities, $10$ of which are chosen to formulate the domain adaptation in our benchmark. 

\noindent\textbf{Sims4Action}~\cite{roitberg2021let} is a synthetic dataset specifically designed for cross-domain evaluation on TSH. It comprises $13,232$ video clips, depicting $10$ daily living activities performed by avatars in the Sims $4$ game, these datasets are used in the Sims4Action $\rightarrow$ TSH adaptation task.
These datasets serve as a robust foundation for exploring DA techniques in various cross-domain scenarios, offering challenges inherent in real-world and synthetic video data.

\subsection{Implementation Details}
We randomly select $N_{shot}$ samples per class on the target domain training set to construct our benchmarks, where $N_{shot} \in \{1,~5,~10,~20\}$ ($1{\sim}20$). 
The few-shot samples are fixed to achieve better knowledge guidance through co-training by using the information provided by the source and target domains. 
To make a fair comparison, all the feature extraction backbone is unified as I3D~\cite{carreira2017quo} initialized with Kinetics400~\cite{kay2017kinetics} pre-trained weight. 
Our model is trained on an NVIDIA-A100 GPU using PyTorch 1.12 with a batch size of $32$, step-wise learning rate decaying at epoch $60$ and $80$, and the Adam optimizer~\cite{kingma2017adam} with an initial learning rate of $0.0001$ for $100$ epochs.
The sliding window size for the feature extraction is set as $N_w = 16$ while temporal zero padding $8$ is used. $K=2$ is chosen for the SDFM. Considering the weights of the losses, $\omega_1 = 0.0001$, $\omega_2 = 1$, $\omega_3 = 1$, $\omega_4=0.01$, and $\omega_5=0.0001$, respectively. $\beta$ in the TRAN-RD is chosen as $0.5$. 
The head's number of RD-MHSA and Scale-wise MHSA is chosen as $8$. 
In SDFM, $\alpha$ is chosen as $0.21$, and $200$ samples are generated for each activity category. The computational complexity of our model is $108.92$ GFLOPS.

\begin{table*}
\caption{Experimental results on the EPIC-KITCHEN~\cite{damen2018scaling} dataset considering six different adaptation settings.}
\vskip-2ex
\label{tab:epic}
\centering
\scalebox{0.69}{    
\begin{tabular}{ll|llll|llll|llll|llll|llll|llll} 
\toprule
\midrule
&\multicolumn{1}{l|}{\multirow{2}{*}{\textbf{Method}}} & \multicolumn{4}{c}{\textbf{\circled{1} D1 $\rightarrow$ D2}} & \multicolumn{4}{c}{\textbf{\circled{2} D2 $\rightarrow$ D1}} & \multicolumn{4}{c}{\textbf{\circled{3} D1 $\rightarrow$ D3 }}
  & \multicolumn{4}{c}{\circled{4} \textbf{D3 $\rightarrow$ D1}} & \multicolumn{4}{c}{\circled{5} \textbf{D2 $\rightarrow$ D3}} & \multicolumn{4}{c}{\circled{6} \textbf{D3 $\rightarrow$ D2}} \\
 &\multicolumn{1}{c|}{} & \multicolumn{1}{c}{\textbf{S-1}} & \multicolumn{1}{c}{\textbf{S-5}} & \multicolumn{1}{c}{\textbf{S-10}} & \multicolumn{1}{c}{\textbf{S-20}} & \multicolumn{1}{c}{\textbf{S-1}} & \multicolumn{1}{c}{\textbf{S-5}} & \multicolumn{1}{c}{\textbf{S-10}} & \multicolumn{1}{c}{\textbf{S-20}} & \multicolumn{1}{c}{\textbf{S-1}} & \multicolumn{1}{c}{\textbf{S-5}} & \multicolumn{1}{c}{\textbf{S-10}} & \multicolumn{1}{c}{\textbf{S-20}} & \multicolumn{1}{c}{\textbf{S-1}} & \multicolumn{1}{c}{\textbf{S-5}} & \multicolumn{1}{c}{\textbf{S-10}} & \multicolumn{1}{c}{\textbf{S-20}} & \multicolumn{1}{c}{\textbf{S-1}} & \multicolumn{1}{c}{\textbf{S-5}} & \multicolumn{1}{c}{\textbf{S-10}} & \multicolumn{1}{c}{\textbf{S-20}} & \multicolumn{1}{c}{\textbf{S-1}} & \multicolumn{1}{c}{\textbf{S-5}} & \multicolumn{1}{c}{\textbf{S-10}} & \multicolumn{1}{c}{\textbf{S-20}} \\ 

\midrule
\multirow{4}{*}{\rotatebox[origin=c]{90}{\textsc{DG}\ }} &Random & \multicolumn{4}{c|}{\textemdash\textemdash\textemdash\textemdash\ 12.5 \textemdash\textemdash\textemdash\textemdash} & \multicolumn{4}{c|}{\textemdash\textemdash\textemdash\textemdash\ 12.3 \textemdash\textemdash\textemdash\textemdash} & \multicolumn{4}{c}{\textemdash\textemdash\textemdash\textemdash\ 12.7 \textemdash\textemdash\textemdash\textemdash} & \multicolumn{4}{c|}{\textemdash\textemdash\textemdash\textemdash\ 12.6 
 \textemdash\textemdash\textemdash\textemdash} & \multicolumn{4}{c|}{\textemdash\textemdash\textemdash\textemdash\ 12.3 
 \textemdash\textemdash\textemdash\textemdash} & \multicolumn{4}{c}{\textemdash\textemdash\textemdash\textemdash\ 12.4 
 \textemdash\textemdash\textemdash\textemdash}\\ 
&KNN & \multicolumn{4}{c|}{\textemdash\textemdash\textemdash\textemdash\ 25.5 
 \textemdash\textemdash\textemdash\textemdash} & \multicolumn{4}{c|}{\textemdash\textemdash\textemdash\textemdash\ 26.9 
 \textemdash\textemdash\textemdash\textemdash} & \multicolumn{4}{c}{\textemdash\textemdash\textemdash\textemdash\ 26.2 
 \textemdash\textemdash\textemdash\textemdash} & \multicolumn{4}{c|}{\textemdash\textemdash\textemdash\textemdash\ 27.4 
 \textemdash\textemdash\textemdash\textemdash} & \multicolumn{4}{c|}{\textemdash\textemdash\textemdash\textemdash\ 28.5 
 \textemdash\textemdash\textemdash\textemdash} & \multicolumn{4}{c}{\textemdash\textemdash\textemdash\textemdash\ 33.9 
 \textemdash\textemdash\textemdash\textemdash}\\ 
&Nearest Center & \multicolumn{4}{c|}{\textemdash\textemdash\textemdash\textemdash\ 24.0 
 \textemdash\textemdash\textemdash\textemdash} & \multicolumn{4}{c|}{\textemdash\textemdash\textemdash\textemdash\ 29.2 
 \textemdash\textemdash\textemdash\textemdash} & \multicolumn{4}{c}{\textemdash\textemdash\textemdash\textemdash\ 29.2 
 \textemdash\textemdash\textemdash\textemdash} & \multicolumn{4}{c|}{\textemdash\textemdash\textemdash\textemdash\ 25.1 
 \textemdash\textemdash\textemdash\textemdash} & \multicolumn{4}{c|}{\textemdash\textemdash\textemdash\textemdash\ 34.0 
 \textemdash\textemdash\textemdash\textemdash} & \multicolumn{4}{c}{\textemdash\textemdash\textemdash\textemdash\ 23.1 
 \textemdash\textemdash\textemdash\textemdash}\\ 
  &Nearest Neighbor & \multicolumn{4}{c|}{\textemdash\textemdash\textemdash\textemdash\ 22.1
 \textemdash\textemdash\textemdash\textemdash} & \multicolumn{4}{c|}{\textemdash\textemdash\textemdash\textemdash\ 24.9 
 \textemdash\textemdash\textemdash\textemdash} & \multicolumn{4}{c}{\textemdash\textemdash\textemdash\textemdash\ 26.6 
 \textemdash\textemdash\textemdash\textemdash} & \multicolumn{4}{c|}{\textemdash\textemdash\textemdash\textemdash\ 26.4
 \textemdash\textemdash\textemdash\textemdash} & \multicolumn{4}{c|}{\textemdash\textemdash\textemdash\textemdash\ 24.1 
 \textemdash\textemdash\textemdash\textemdash} & \multicolumn{4}{c}{\textemdash\textemdash\textemdash\textemdash\ 28.7 
 \textemdash\textemdash\textemdash\textemdash}\\
\midrule
\multirow{10}{*}{\rotatebox[origin=c]{90}{\textsc{FSDA-AR}}} &CoMix~\cite{sahoo2021contrast}  & 31.5 & 32.0 & 34.3 & 35.7& 25.2 & 27.9 & 31.1 & 29.9 & 30.0 & 30.4 & 28.5 & 30.8 & 30.4 & 30.2 & 28.8 & 30.4 & 34.0 & 34.5 & 35.3 & 34.4 & 36.0 & 35.7 & 34.5 & 35.1\\
&CO$^2$A~\cite{da2022dual} & 31.7 & 33.5 & 33.6 & 33.3 & 32.6 & 38.4 & 36.1 & 39.8 & 30.2 &36.4& 38.3 & 38.0 & 34.0 & 36.6 & 40.5 & 40.0 & 34.7 & 36.6 & 40.5 & 40.0 & 32.6 & 36.8 & 39.0 & 38.6\\
&TA$^3$N~\cite{chen2019taaan} & 36.8 & 39.0 & 40.2 & 43.5 & 36.8 & 38.9 & 40.4 & 40.5 & 36.7 & 40.2 & 41.0 & 40.3& 33.1 & 40.0 & 40.9 & 41.8 & 41.1 & 43.4 & 43.5 & 45.6 & 42.8 & 45.8 & 46.5 & 46.5 \\ %

&TranSVAE~\cite{wei2022transvae} & 32.9 & 39.5 & 39.5 & 42.8 & 35.3 & 40.4 & 37.5 & 41.7 &  37.0& 39.1 & 40.3 & 42.3& 36.1 & 38.2 & 37.5 & 41.4 & 42.8 &  44.9 & 44.5 & 45.9 & 41.2 & 44.4 & 45.6 & 45.6\\ %
\cmidrule{2-26}
& TRX~\cite{perrett2021temporal} & 24.8 & 25.0 & 25.2 & 25.9 & 26.1 & 27.7 & 30.7 & 31.6 & 25.3 & 25.9 & 28.1 & 28.8 & 26.6 & 28.9 & 29.3  & 30.0 & 28.4 & 28.0 & 30.6 & 31.9 & 28.8 & 29.1 & 28.4 & 33.1\\
& HyRSM~\cite{wang2022hybrid} & 31.1 & 33.5 & 34.0 & 37.2 & 33.4 & 32.7 & 33.9 & 34.8 & 33.2 & 37.2 & 36.5 & 36.7 & 35.0 & 34.8 & 35.7 & 35.0 & 40.4 & 40.3 & 41.2 & 41.4 & 41.6 & 41.8 & 41.5 & 41.5\\
\cmidrule{2-26}
&FS-ADA~\cite{li2021supervised} & 36.4 & 38.1 & 38.4 & 37.7 &  34.7&  36.8 & 39.1 & 39.3 & 36.1 & 37.4 & 38.2 &  40.5&  32.2 & 39.8 & 35.9 & 38.6 & 42.4 & 42.5 & 42.0 & 44.3 & 40.4 & 43.5 & 42.1 &  42.2\\ %
&PASTN~\cite{gao2020pairwise} & 33.3 & 38.2 & 37.7 & 41.3 & 34.0 & 38.9 & 36.8 & 40.9 & 35.3 & 39.4 & 39.0 & 41.0&   33.6 & 37.9 & 38.2 & 41.1& 39.2 & 43.1 & 44.4 & 44.6  & 43.0 & 45.5 & 45.9 & 45.8 \\ %
& SSA$^2$lign~\cite{xu2023augmenting} & 32.0 & 40.4 & 37.6 & 41.5 & 31.3 & 40.1 & 40.5 & 41.6 & 30.1 & 39.3 & 42.0 & 42.6 & 34.5 & 38.9 & 41.1 & 39.1 & 28.7 & 42.9 & 42.1 & 44.5 & 32.3 & 38.7 & 41.9 & 42.7 \\
\cmidrule{2-26}
\rowcolor{gray!15}& RelaMiX (ours) & \textbf{39.1} & \textbf{43.9} & \textbf{43.7} & \textbf{47.9} & \textbf{38.4}& \textbf{41.6} & \textbf{42.1} & \textbf{42.8} & \textbf{38.4}& \textbf{42.1} & \textbf{42.5} & \textbf{43.1} & \textbf{37.9} & \textbf{41.6} & \textbf{42.3} & \textbf{42.5} & \textbf{45.1} & \textbf{46.2} & \textbf{47.4} & \textbf{46.5} & \textbf{45.5} & \textbf{48.0} & \textbf{48.1} & \textbf{48.1} \\
\hline
\bottomrule
\end{tabular}}
\vskip-1ex
\end{table*}

\subsection{Analysis of the Benchmark}

The performance results for the transfer tasks involving the UCF~\cite{soomro2012ucf101} to HMDB~\cite{kuehne2011hmdb}, HMDB~\cite{kuehne2011hmdb} to UCF~\cite{soomro2012ucf101}, EPIC-KITCHEN~\cite{damen2018scaling}, and Sims4Action~\cite{roitberg2021let} to ToyotaSmartHome~\cite{das2019toyota} (TSH) settings are presented in Table~\ref{tab:hmdb} (\circled{1}, \circled{2}, \circled{3}, and \circled{4}), respectively. We also provide per-split performances on the EPIC-KITCHEN in Table~\ref{tab:epic}, where S-1, S-5, S-10, and S-20 indicate shot 1, shot 5, shot 10, and shot 20.
Both UDA methods, implemented within the context of the FSDA-AR task, as well as previously published FSDA-AR techniques, show notable performance, consistently outperforming random baseline benchmarks. 
This observation underscores the effective reduction of domain gaps when utilizing a limited number of labeled shots. 
Notably, the reduction in the required number of target domain samples is substantial when we compare FSDA-AR with the well-established UDA task. 
For instance, in a scenario with only $5$ labeled shots, the total required sample count is approximately only $1.6\%$ of that used in conventional UDA setting on EPIC-KITCHEN D1$\rightarrow$D2.
However, it is noteworthy that the TranSVAE method~\cite{wei2022transvae} exhibits inferior performance in the FSDA-AR task, even though it outperforms TA$^3$N~\cite{chen2019taaan} in the context of UDA task, as reported in Table~\ref{tab:comparison_epic_kitchen}.

This discrepancy suggests that the disentanglement method applied in the context of domain adaptation relies heavily on the availability of large-scale data from the target domain to effectively capture adaptation cues.
Similar trends can be observed in the case of the CoMix approach~\cite{sahoo2021contrast}, which relies on the diversity of backgrounds in the target domain to achieve adaptation, particularly in scenarios involving datasets with substantial domain gaps.
In all our experiments, we employ the I3D backbone~\cite{carreira2017quo} to ensure equitable comparisons. 
To facilitate an assessment relative to the SSA$^2$Lign approach~\cite{xu2023augmenting}, we replace the TimesFormer~\cite{bertasius2021spacetime} backbone in SSA$^2$Lign with the I3D~\cite{carreira2017quo} backbone. 
The reduction in performance observed in the adapted SSA$^2$Lign model when applied to the leveraged datasets can be attributed to two primary factors. 

Firstly, it stems from disparities in domain gaps in our experimental configurations. Secondly, it arises from SSA$^2$Lign's reliance on transformer features sourced from the TimesFormer~\cite{bertasius2021spacetime} backbone.
However, to ensure an equitable comparison with methods tailored originally for UDA, it is imperative to standardize the backbone as I3D~\cite{carreira2017quo}. The rationale behind this standardization is that I3D is commonly employed in UDA works~\cite{wei2022transvae}, thereby allowing us to effectively demonstrate that the observed performance enhancements are a consequence of our proposed method, rather than a consequence of backbone substitution.
To maximize the utility of the limited target-domain samples provided, we take into account three crucial elements concerning the latent space source domain feature generation, the generalizability of the temporal aggregator considering multi-scale relationships, and the inter-domain alignment by using target-domain prototypes.

\begin{table}[t]
\caption{Task comparison between FSDA-AR and UDA.}
\vskip-2ex
\label{tab:comparison_epic_kitchen}
\newcommand{\emdashnum}{7}
\centering
\scalebox{0.9}{\begin{tabular}{l|llll} 
\toprule
\midrule
\multicolumn{5}{l}{\textbf{\circled{1} UDA approaches on UDA task on EPIC-KITCHEN (2645 shots).}} \\ 
TranSVAE~\cite{wei2022transvae} & \multicolumn{4}{l}{\remdash{\emdashnum}\ 52.6 \remdash{\emdashnum}}\\ 
CoMix~\cite{sahoo2021contrast} & \multicolumn{4}{l}{\remdash{\emdashnum}\ 43.2
 \remdash{\emdashnum}}  \\ 
TA$^3$N~\cite{chen2019taaan}& \multicolumn{4}{l}{\remdash{\emdashnum}\  39.9
 \remdash{\emdashnum}}  \\ 
DANN~\cite{ganin2015unsupervised} & \multicolumn{4}{l}{\remdash{\emdashnum}\ 39.2  
 \remdash{\emdashnum}}  \\ 
ADDA~\cite{tzeng2017adversarial} & \multicolumn{4}{l}{\remdash{\emdashnum}\ 39.2  
 \remdash{\emdashnum}}  \\ 
\midrule
\multicolumn{5}{l}{\textbf{\circled{2} TA$^3$N and RelaMiX on FSDA-AR task on EPIC-KITCHEN.}} \\ 
\midrule
\textbf{Method} & \textbf{Shot-1} & \textbf{Shot-5} & \textbf{Shot-10} & \textbf{Shot-20} \\ 
TA$^3$N~\cite{chen2019taaan} & 38.9 & 41.8 & 42.1 & 43.0 \\
 \rowcolor{gray!15} RelaMiX (ours) & 41.0 & 44.4 & 44.5 & 45.1\\ 
 \midrule
\multicolumn{5}{l}{\textbf{\circled{3} UDA approaches for UDA tasks on UCF $\rightarrow$ HMDB (840 shots).}} \\ 
TranSVAE~\cite{wei2022transvae} & \multicolumn{4}{l}{\remdash{\emdashnum}\ 87.8 \remdash{\emdashnum}}\\ 
CoMix~\cite{sahoo2021contrast} & \multicolumn{4}{l}{\remdash{\emdashnum}\ 86.7
 \remdash{\emdashnum}}  \\ 
TA$^3$N~\cite{chen2019taaan}& \multicolumn{4}{l}{\remdash{\emdashnum}\  81.4
 \remdash{\emdashnum}}  \\ 
DANN~\cite{ganin2015unsupervised} & \multicolumn{4}{l}{\remdash{\emdashnum}\ 80.1  
 \remdash{\emdashnum}}  \\ 
ADDA~\cite{tzeng2017adversarial} & \multicolumn{4}{l}{\remdash{\emdashnum}\ 79.2  
 \remdash{\emdashnum}}  \\ 
\midrule
\multicolumn{5}{l}{\textbf{\circled{4} TA$^3$N and RelaMiX for FSDA-AR task on UCF $\rightarrow$ HMDB.}} \\ 
\midrule
\textbf{Method} & \textbf{Shot-1} & \textbf{Shot-5} & \textbf{Shot-10} & \textbf{Shot-20} \\ 
TA$^3$N ~\cite{chen2019taaan} & 83.3 & 88.9 & 88.3 & 91.7\\ 
\rowcolor{gray!15} RelaMiX (ours) & 84.4 & 89.7 & 90.3 &  92.8\\ 
 \midrule
 \multicolumn{5}{l}{\textbf{\circled{5} UDA approaches for UDA tasks on HMDB $\rightarrow$ UCF (1438 shots).}} \\ 
TranSVAE~\cite{wei2022transvae} & \multicolumn{4}{l}{\remdash{\emdashnum}\ 99.0 \remdash{\emdashnum}}\\ 
CoMix~\cite{sahoo2021contrast} & \multicolumn{4}{l}{\remdash{\emdashnum}\ 93.9
 \remdash{\emdashnum}}  \\ 
TA$^3$N~\cite{chen2019taaan}& \multicolumn{4}{l}{\remdash{\emdashnum}\  90.5
 \remdash{\emdashnum}}  \\ 
DANN~\cite{ganin2015unsupervised} & \multicolumn{4}{l}{\remdash{\emdashnum}\ 88.1  
 \remdash{\emdashnum}}  \\ 
ADDA~\cite{tzeng2017adversarial} & \multicolumn{4}{l}{\remdash{\emdashnum}\ 88.4  
 \remdash{\emdashnum}}  \\ 
\midrule
\multicolumn{5}{l}{\textbf{\circled{6} TA$^3$N and RelaMiX for FSDA-AR task on HMDB $\rightarrow$ UCF.}} \\ 
\midrule
\textbf{Method} & \textbf{Shot-1} & \textbf{Shot-5} & \textbf{Shot-10} & \textbf{Shot-20} \\ 
TA$^3$N~\cite{chen2019taaan} & 93.7 & 95.1 & 97.5 & 98.0 \\
\rowcolor{gray!15} RelaMiX (ours) & 95.6 & 96.5 & 97.7 &  98.2\\ 
  \midrule
 \multicolumn{5}{l}{\textbf{\circled{7} UDA approaches for UDA task on Sims4Action $\rightarrow$ TSH (8552 shots).}} \\ 
%
Schneider \textit{et al.}~\cite{schneider2022pose}  & \multicolumn{4}{l}{\remdash{\emdashnum}\ 36.3 
 \remdash{\emdashnum}}  \\ 
TA$^3$N~\cite{chen2019taaan} (\cite{schneider2022pose})& \multicolumn{4}{l}{\remdash{\emdashnum}\ \ \ 8.0
 \remdash{\emdashnum}}  \\ 
\midrule
\multicolumn{5}{l}{\textbf{\circled{8} TA$^3$N and RelaMiX for FSDA-AR task on Sims4Action $\rightarrow$ TSH.}} \\
\midrule
\textbf{Method} & \textbf{Shot-1} & \textbf{Shot-5} & \textbf{Shot-10} & \textbf{Shot-20} \\ 
TA$^3$N~\cite{chen2019taaan} & 21.1 & 29.8 & 35.8 &  42.7\\
\rowcolor{gray!15} RelaMiX (ours) & 24.6 & 31.4 & 36.7 &  45.7\\ 
\midrule
\bottomrule
\end{tabular}}
\vskip-1ex
\end{table}
\begin{figure}[t!]
\centering
\includegraphics[width=0.99\linewidth]{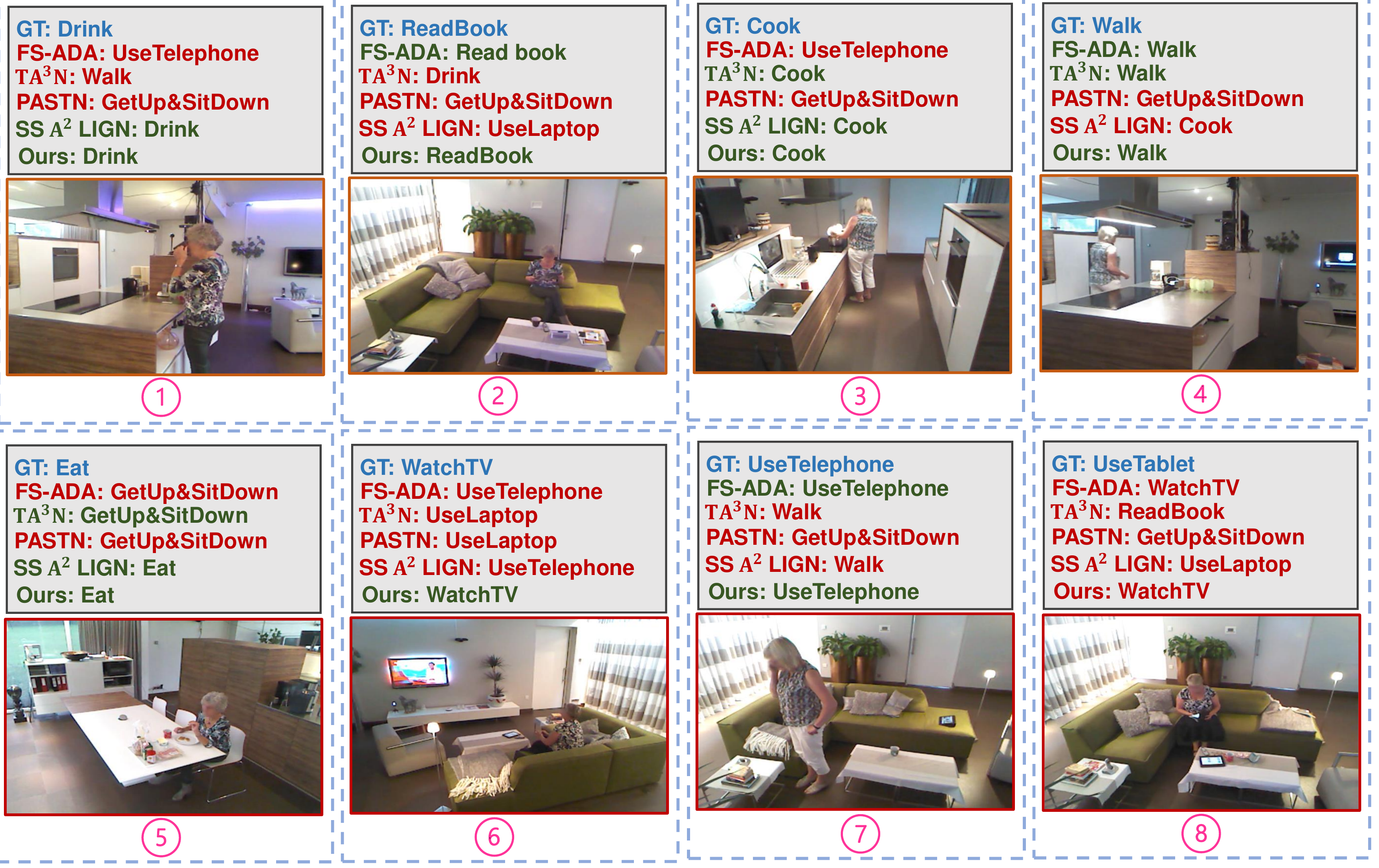}
\vskip-1ex
\caption{Qualitative results for FSDA-AR on Shot-$20$ Sims4Action~\cite{roitberg2021let} $\rightarrow$ TSH~\cite{das2019toyota}.}
\vskip-2ex
\label{fig:qualitative}
\end{figure}

Comparing against the best baseline for each shot-setting, our proposed RelaMiX achieves $2.8\%$, $2.7\%$, $2.3\%$, and $1.9\%$ performance improvements for $1{\sim}20$-shot setting on EPIC-KITCHEN~\cite{damen2018scaling} in terms of FSDA-AR in Table~\ref{tab:hmdb} \circled{3}. The per-split performances are showcased in Table~\ref{tab:epic}, where RelaMiX shows consistent superior results on the FSDA-AR task for each split of EPIC-KITCHEN~\cite{damen2018scaling}. Some UDA approaches implemented into FSDA-AR do not show better performances compared with statistic baselines, indicating that fewer target domain samples may cause overfitting to some approaches that require a large number of samples, especially on the datasets with large domain differences, \eg, CoMix~\cite{sahoo2021contrast} on EPIC-KITCHEN~\cite{damen2018scaling}.

Regarding the FSDA-AR task on the UCF~\cite{soomro2012ucf101} $\rightarrow$ HMDB~\cite{kuehne2011hmdb} and HMDB~\cite{kuehne2011hmdb} $\rightarrow$ UCF~\cite{soomro2012ucf101} introduced in Table~\ref{tab:hmdb} \circled{1} and \circled{2}, all the approaches show promising performance even under $1$-shot.
When $N_{shot}\geq5$ on the UCF~\cite{soomro2012ucf101} $\rightarrow$ HMDB~\cite{kuehne2011hmdb}, CoMix~\cite{sahoo2021contrast}, CO$^2$A~\cite{da2022dual}, and TA$^3$N~\cite{chen2019taaan} under FSDA-AR are better than the state-of-the-art performance $87.8\%$ delivered by TranSVAE for UDA task as in Table~\ref{tab:comparison_epic_kitchen} \circled{1}, which demonstrates that FSDA-AR is more efficient compared with UDA when facing with small domain gap.
Compared with the approach with the best performance among all the baselines for $1{\sim}20$ shot settings, RelaMiX achieves performance improvements with $1.7\%$, $2.2\%$, $1.4\%$, and $0.5\%$ for FSDA-AR on the UCF~\cite{soomro2012ucf101} $\rightarrow$ HMDB~\cite{kuehne2011hmdb} and $0.4\%$, $2.1\%$, $0.4\%$, and $0.4\%$ on the HMDB~\cite{kuehne2011hmdb} $\rightarrow$ UCF~\cite{soomro2012ucf101}, respectively.
Compared with the baseline with the best performance for $1$-$20$ shot settings on Sims4Action~\cite{roitberg2021let} $\rightarrow$ TSH~\cite{das2019toyota}, RelaMiX achieves performance improvements with $2.8\%$, $1.2\%$, $3.1\%$, and $6.5\%$, introduced in Table~\ref{tab:epic}~\circled{4}.
The consistent performance enhancements produced by RelaMiX across various datasets indicate that the proposed method effectively utilizes the guidance provided by the few-shot labeled samples from the target domain.
Furthermore, RelaMiX can achieve a generalizable temporal aggregation that accounts for diverse domain differences.
In terms of video-based domain adaptation, the FSDA-AR task exhibits comparable performance to the UDA task for human activity recognition, as reported in Table~\ref{tab:comparison_epic_kitchen}. 
Consequently, our experiments confirm the feasibility of FSDA-AR and we believe it to be an essential future research direction in domain adaptation for human activity recognition.

\subsection{Ablation Studies}

\begin{table}[t]
\caption{Module ablation on EPIC-KITCHEN~\cite{damen2018scaling} D1 $\rightarrow$ D2.}
\vskip-2ex
\label{tab:module_ablation}
\centering
\begin{tabular}{lllll}
\toprule
\midrule
\textbf{Method}   & \textbf{Shot-1} & \textbf{Shot-5} & \textbf{Shot-10} & \textbf{Shot-20} \\
\midrule
w/o TARD-RD &    36.3    &    40.3    &     40.9    &         44.1 \\
w/o CDIA &    37.9    &     41.3   &    40.7     &     47.2    \\
w/o SDFM &    34.7    &     42.5   &     42.3    &     45.3    \\
w/ All   &    \textbf{39.1}    &   \textbf{43.9}     &   \textbf{43.7}      &      \textbf{47.9}    \\
\midrule
\bottomrule
\end{tabular}
\vskip-2ex
\end{table}

To assess the efficacy of each proposed mechanism, we conduct ablation experiments on the EPIC-KITCHEN~\cite{damen2018scaling} dataset, specifically the D1 $\rightarrow$ D2 split, as detailed in Table~\ref{tab:module_ablation}. 
Initially, we compare RelaMiX against RelaMiX w/o TARD-RD, employing Temporal Relation Networks (TRN) as an alternative for temporal aggregation. 
The results indicate that RelaMiX outperforms RelaMiX w/o TARD-RD, achieving performance improvements of $2.8\%$, $3.6\%$, $2.8\%$, and $3.8\%$ across $1{\sim}20$ shot settings. 
These findings highlight the superiority of our TRAN-RD method for temporal aggregation in the FSDA-AR task. 
The integration of relational attention with relation dropout and scale-wise self-attention is verified effective in facilitating generalizable temporal aggregation and feature learning.

Subsequently, we compare RelaMiX with RelaMiX w/o the SDFM component. 
In this scenario, RelaMiX demonstrates superiority over its SDFM-lacking counterpart, exhibiting performance improvements of $4.4\%$, $1.4\%$, $1.4\%$, and $2.6\%$ for the $1{\sim}20$ shot settings. 
These results indicate that SDFM effectively enhances the learned embeddings of the target domain.
Finally, a comparison is made between RelaMiX and RelaMiX w/o CDIA. 
RelaMiX outperforms the CDIA-lacking variant by $1.2\%$, $2.6\%$, $3.0\%$, and $0.7\%$ for the $1{\sim}20$ shot settings, revealing that CDIA plays a significant role in bridging the domain gap by extracting pertinent information few target-domain samples. More ablations can be found in the supplementary.

\subsection{Analysis of Qualitative Results}
Apart from the quantitative analysis, we further assess qualitative results from the proposed FSDA-AR task. 
As shown in \circledmagenta{1}--\circledmagenta{8} in Figure~\ref{fig:qualitative}, we visualize ten results of activity recognition from the setting of $20$-shot Sims4Action $\rightarrow$ TSH, including all of the ten classes leveraged for FSDA-AR. 
For each sample, we compare the activity predictions from FS-ADA~\cite{li2021supervised}, TA$^3$N~\cite{chen2019taaan}, SSA$^2$Lign~\cite{xu2023augmenting}, PASTN~\cite{gao2020pairwise}, and our RelaMiX, respectively. 
Since the domain gap between the synthesized dataset and the real dataset is large, most of the leveraged baselines can not guarantee superior performances. 
The RelaMiX, considering temporal generalizability, latent space diversity, and cross-domain alignment, boasts impressive performance in challenging settings and showcases that the novel techniques significantly enhance temporal aggregation generalizability.

\section{Conclusion}
This study delves into video-based human activity recognition through Few-Shot Domain Adaptation (FSDA-AR), introducing a new benchmark with five challenging datasets covering real-world, egocentric, and synthetic videos. We present RelaMiX, featuring a Temporal Relational Attention Network with Relation Dropout, a Statistical Distribution-Based Feature Mixture method, and Cross-Domain Information Alignment, achieving top performance across diverse FSDA-AR scenarios. RelaMiX's effectiveness, comparable to the UDA task with far fewer target domain samples, highlights FSDA-AR's potential for improving data efficiency in domain adaptation, suggesting a valuable research direction.

\bibliographystyle{ACM-Reference-Format}
\bibliography{bib}

\appendix
\section{Social Impact and Limitations}
\noindent\textbf{Societal Impacts.}
In this work, we first construct the few-shot domain adaptation for activity recognition benchmark to enable more future research since the benchmark of the published paper~\cite{gao2020pairwise} is unfortunately unavailable. We constructed the FSDA-AR benchmarks on five established activity recognition datasets encompassing more diverse and challenging domains, \eg, HMDB-51~\cite{kuehne2011hmdb}, UCF-101~\cite{soomro2012ucf101}, EPIC-KITCHEN~\cite{damen2018scaling}, Sims4Action~\cite{roitberg2021let} and Toyota Smart Home (TSH)~\cite{das2019toyota}. We further proposed a new approach, \ie, RelaMiX, which shows state-of-the-art performances on the proposed benchmark. Through our experiments we find that all the leveraged approaches show comparable performances on the FSDA-AR task compared with UDA for human activity recognition. Using very few yet labeled samples sometimes even shows better performance compared with using a large-scale unlabeled training set from the target domain, which emphasizes a new solution in the application while dealing with the video-based domain adaptation for a small domain gap. Unlike pixel-wise annotation which is used for semantic segmentation task, less time is required for the labeling work of human activity recognition tasks. In this case, FSDA-AR will be piratically useful since it will reduce $70\%$ to $98\%$ training samples on the target domain. However, our method does have false predictions as illustrated in the qualitative analysis in the main paper, which means our model has the potential to give offensive predictions, misclassification, and biased content which may cause false predictions resulting in a negative social impact.

\noindent\textbf{Limitations.} In order to make a fair comparison with the existing UDA approaches on the UDA task, we unified the feature extraction backbone of all the leveraged approaches as I3D~\cite{carreira2017quo} pretrained on the Kinetics400~\cite{kay2017kinetics} dataset. However, there is also other existing frameworks that could serve as the feature extractor which will be considered in future work. The performance of the RelaMiX approach still has a large space for further improvement on the FSDA-AR benchmark and RelaMiX may cause false prediction in the real-world application as shown in the qualitative analysis part in our main paper. We only tackle the FSDA-AR for RGB video. Multi-modality FSDA-AR is not involved in this work, which will be addressed in future work.

\section{Analysis of the t-SNE Visualization}
\begin{figure}[htb]
\centering
\includegraphics[width=0.94\linewidth]{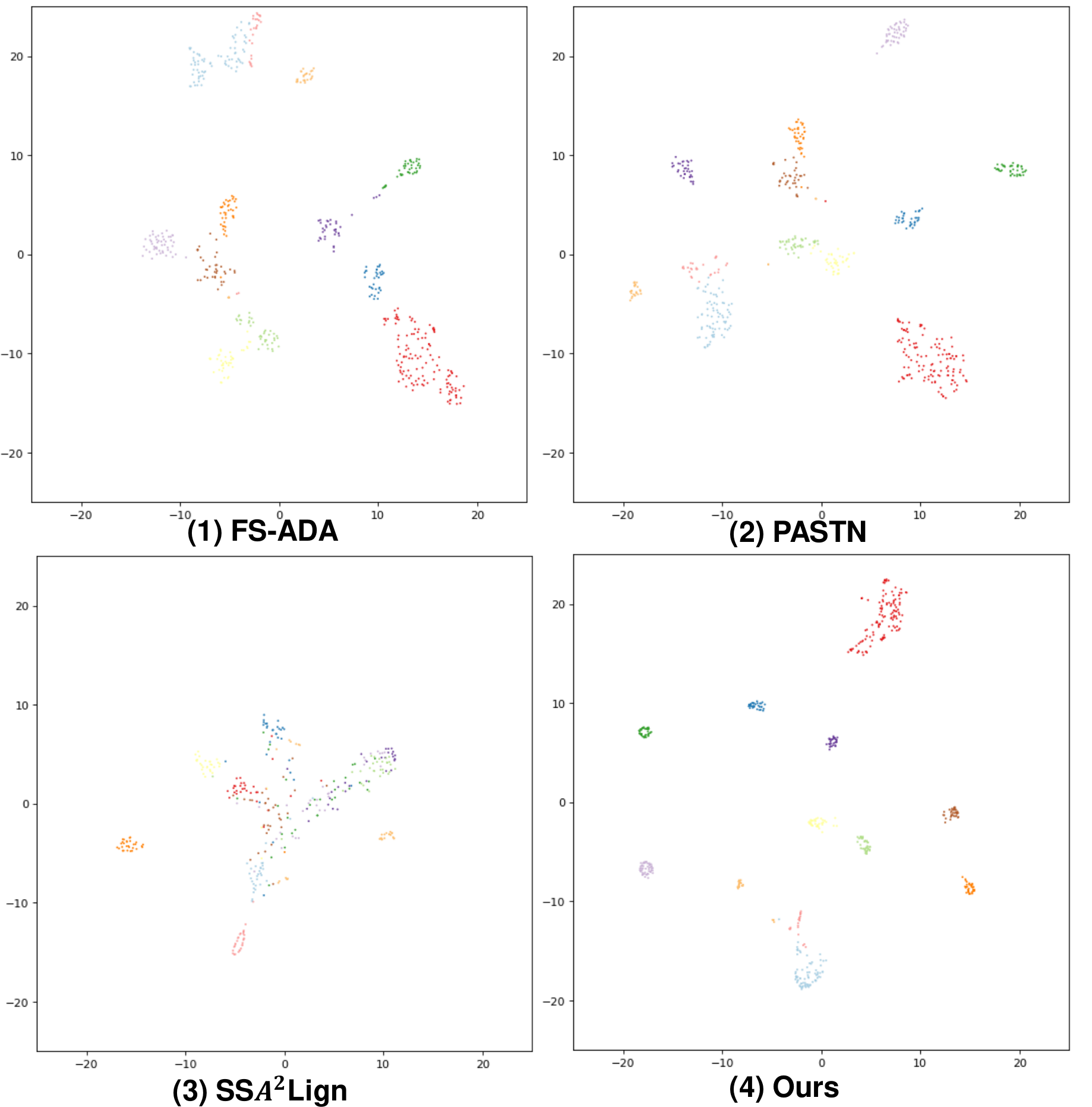}
\caption{The t-SNE feature visualization~\cite{van2008visualizing} on the UCF test set~\cite{soomro2012ucf101} for FSDA-AR on $20-$Shot HMDB~\cite{kuehne2011hmdb} $\rightarrow$ UCF~\cite{soomro2012ucf101}.}
\label{fig:tsne}
\end{figure}

To investigate the performance of few-shot domain adaptation on latent space, t-SNE distribution~\cite{van2008visualizing} is presented in Figure~\ref{fig:tsne}. From (a) to (d) are the t-SNE distributions from FS-ADA, PASTN, TA$^3$N, and our RelaMiX. 
Under the setting of HMDB~\cite{kuehne2011hmdb} $\rightarrow$ UCF~\cite{soomro2012ucf101}, the samples shown in Figure~\ref{fig:tsne} are selected from the UCF101 test set. 
Compared to the other methods, the features from our RelaMiX are more distinguishable across different classes in the latent space, showing a better generalization ability of RelaMiX, which is crucial in FSDA-AR.

\begin{table}[htb]
\caption{Module ablation for TRAN-RD mechanism on EPIC-KITCHEN~\cite{damen2018scaling} D1 $\rightarrow$ D2 considering different shots from the target domain.}
\label{tab:tranrd_ablation}
\centering
\begin{tabular}{lllll}
\toprule
\midrule
\textbf{Method}   & \textbf{S-1} & \textbf{S-5} & \textbf{S-10} & \textbf{S-20} \\
\midrule
 w/o RD-MHSA &  31.6  &   37.1  &   39.2  &  39.2   \\
 w/o Scale-wise MHSA &  28.1   &  37.9   &   39.3    &   39.3   \\
 w/o RD &    32.3 &   39.5  &  37.3    &  39.7   \\
w/ All   &    \textbf{39.1}    &   \textbf{43.9}     &   \textbf{43.7}      &      \textbf{47.9}    \\
\midrule
\bottomrule
\end{tabular}
\end{table}

\begin{table}[htb]
\caption{Module ablation for CDIA on EPIC-KITCHEN~\cite{damen2018scaling} D1 $\rightarrow$ D2 considering different shots from the target domain.}
\label{tab:cdia_ablation}
\centering
\begin{tabular}{lllll}
\toprule
\midrule
\textbf{Method}   & \textbf{S-1} & \textbf{S-5} & \textbf{S-10} & \textbf{S-20} \\
\midrule
 w/o prototypes &  31.5   &   35.7  &   35.7   &    42.5  \\
 w/o mixed domain negatives &  34.4   &  38.3   &     37.2  &   39.3   \\
w/ All   &    \textbf{39.1}    &   \textbf{43.9}     &   \textbf{43.7}      &      \textbf{47.9}    \\
\midrule
\bottomrule
\end{tabular}
\end{table}

\begin{table}[htb]
\caption{Module ablation for SDFM on EPIC-KITCHEN~\cite{damen2018scaling} D1 $\rightarrow$ D2 considering different shots from the target domain.}
\label{tab:sfdm_ablation}
\centering
\begin{tabular}{lllll}
\toprule
\midrule
\textbf{Method}   & \textbf{S-1} & \textbf{S-5} & \textbf{S-10} & \textbf{S-20} \\
\midrule
 K=1 &   33.6  &  38.4   &    39.1  &  44.5    \\
 K=3 &  32.7   &   38.0  &   38.5    &  43.6    \\
 K=4 &   34.9  &   39.3  &    38.8   &    43.9  \\
 \midrule
K=2 &    \textbf{39.1}    &   \textbf{43.9}     &   \textbf{43.7}      &      \textbf{47.9}    \\
\midrule
\bottomrule
\end{tabular}
\end{table}

\begin{table}[htb]
\caption{Comparison between TRAN-RD with other temporal aggregation methods on EPIC-KITCHEN~\cite{damen2018scaling} D1 $\rightarrow$ D2 considering different shots from the target domain. The GFLOPS are provided, while all the methods share the same feature extractor I3D~\cite{carreira2017quo} with $108$ GFLOPS.}
\label{tab:temporal_ablation}
\centering
\begin{tabular}{l|l|llll}
\toprule
\midrule
\textbf{Method}  &\textbf{GFLOPS} & \textbf{S-1} & \textbf{S-5} & \textbf{S-10} & \textbf{S-20} \\
\midrule
 LSTM & 0.09 & 31.3  &   40.9  &  35.2    &   37.7   \\
 GRU & 0.10 &32.5 &  32.9   &  36.5     &  39.1  \\
 TRN & 0.04 &33.2   &   43.1  &   40.1    &   41.5   \\
 \midrule
TRAN-RD & 0.92  & \textbf{39.1}    &   \textbf{43.9}     &   \textbf{43.7}      &      \textbf{47.9}    \\
\midrule
\bottomrule
\end{tabular}
\end{table}
\section{Comparison with Other Domain Adaptation Settings.}
We begin by distinguishing the few-shot domain adaptation task from other domain adaptation settings commonly used in previous research.
The task we address is Few-Shot Domain Adaptation (FSDA), where a very small amount of labeled examples are available for each category in the target domain. The comparison among the commonly addressed tasks of Semi-Supervised Domain Adaptation (SSDA)~\cite{saito2019semi, yang2021deep, yoon2022semi, Li_2021_CVPR}, Unsupervised Domain Adaptation (UDA)~\cite{chen2019taaan, wei2022transvae, sahoo2021contrast, kang2019contrastive,choi2020shuffle, xu2022learning}, and our FSDA is provided in Fig.1 in our main paper.
In contrast to FSDA, both UDA and SSDA necessitate a large amount of unlabeled data in the target domain to construct the training set, while collecting such large-scale data with high quality might not be feasible in certain applications. FSDA, on the other hand, requires only a small number of labeled examples from the target domain, balancing the trade-off between data collection expenses and labeling efforts. 
In our benchmark, the leveraged few-shot setting results in a $70\%$ to $98\%$ reduction of data from the target domain to construct the training set compared to SSDA and UDA. Activity recognition does not rely on pixel-level dense annotation, which makes the trade-off between annotation and data collection considerable.
It is crucial to note that FSDA is distinct from domain adaptation for few-shot learning~\cite{cong2021inductive, zhang2021knowledge}, which is specifically designed for adapting to new classes with limited examples, while we are adapting to new domains. 
FSDA does not encompass new activity classes.
\section{More Ablations}
\subsection{Ablation of the TRAN-RD.} We deliver the ablation experiments of the TRAN-RD in Table~\ref{tab:tranrd_ablation}, where \textit{w/o RD-MHSA} means that the RD-MHSA is replaced by multi-layer perceptron (MLP), \textit{w/o Scale-wise MHSA} means that we use mean average to achieve multi-scale aggregation, and \textit{w/o RD} indicates that the relation dropout is discarded. First, when we compare TRAN-RD with the variant \textit{w/o RD-MHSA}, we find out that by using RD-MHSA, the model can achieve $7.5\%$, $6.8\%$, $4.5\%$, and $8.7\%$ performance improvements for FSDA-AR considering $1{\sim}20$ shot settings, which showcases the importance of the RD-MHSA on snippet-wise temporal information aggregation in our RelaMiX.
Then, we compare TRAN-RD with the ablation \textit{w/o Scale-wise MHSA}, we observe that Scale-wise MHSA can achieve performance gains of $11.0\%$, $6.0\%$, $4.4\%$, and $8.6\%$ for $1{\sim}20$ shot settings, indicating the superiority of the scale-wise information reasoning ability of this superior design.
Finally, to showcase the importance of the relation dropout design, we compare TRAN-RD with \textit{w/o RD}, where the relation dropout brings a performance boost by $6.8\%$, $4.4\%$, $6.4\%$, and $8.2\%$ for the $1{\sim}20$ shot settings, respectively. This ablation study demonstrates the collaboration of each part of the module design helps to obtain a generalizable temporal aggregator for the FSDA-AR setting.

\subsection{Ablation of the CDIA.}
We conduct ablation experiments in Table~\ref{tab:cdia_ablation} to demonstrate the reason behind the supervision design. Two ablation experiments are executed for CDIA, which are \textit{w/o prototypes} by replacing the prototype-based positive anchors with the randomly temporal permuted anchor embedding, and \textit{w/o mixed domain negatives} by replacing the mixed domain negatives with the source-domain negative anchor embeddings.
Compared with variant \textit{w/o prototypes}, using the target domain prototypes as the positive anchors achieves performance improvements of $7.6\%$, $8.2\%$, $8.0\%$, and $5.4\%$ for the $1{\sim}20$ shot settings. 

Compared with \textit{w/o mixed domain negatives}, using the domain negatives achieves performance improvements of $4.7\%$, $5.6\%$, $6.5\%$, and $8.6\%$ for the $1{\sim}20$ shot settings. The aforementioned observations turn out that mixed domain negatives and target domain prototypes collaborate together to contribute a more reasonable FSDA-AR supervision when only few-shot target domain samples are available.

\subsection{Ablation of the SDFM.}
We conduct the ablation study for SFDM in Table~\ref{tab:sfdm_ablation} to observe the influence brought by different numbers of cluster centers, where the experiments are done for $K\in \{1,2,3,4\}$. We found the setting $K=2$ generally works well for the feature mixture.

\subsection{Comparison with Other Temporal Aggregators.}
We further conduct a comparison study among TRAN-RD and other existing temporal aggregators, \ie, LSTM, GRU, and TRN~\cite{zhou2018temporal}, in Table~\ref{tab:temporal_ablation}.
Our TRAN-RD outperforms all the others by a large margin for the $1{\sim}20$ shot settings. TRN can achieve a good performance on the $5{\sim}20$ shot settings, however, it can not achieve a generalizable temporal aggregation by using an extremely small shot number, \eg, on $1$ shot setting, which is an important setting when the data in the target domain is extremely hard to acquire.
The other temporal aggregators, \eg, LSTM and GRU, also have the aforementioned problem.

Our proposed TRAN-RD overcomes this difficulty by using superior relation-based techniques to achieve temporal aggregation in different scale settings. Alongside the recognition performances, we also provide GFLOPS of different temporal aggregators to illustrate the computational difficulties of our approach during inference. 
Since all the baselines chosen in our benchmark are unified to use I3D~\cite{carreira2017quo} as the feature extractor, which takes the most of the computational complexity, \ie, $108$ GFLOPS, directly comparing the GFLOPS of the temporal aggregator will be more obvious.
Our TRAN-RD brings a computational complexity increase by $0.92$ GFLOPS due to the leverage of the RD-MHSA and Scale-wise MHSA mechanism. However, this increase is small compared with the GFLOPS of the video backbone I3D~\cite{carreira2017quo} with $108$ GFLOPS.
Moreover, CDIA and SFDM only participate in the training phase and they do not contribute to the computational complexity during the inference time.
\subsection{Analysis of the Target Domain Sample Number of FSDA-AR and UDA}
We present the required sample number to formulate the training set on the target domain separately for FSDA-AR and UDA tasks on all the leveraged DA settings in Table~\ref{tab:details}.
Compared with UDA, FSDA-AR requires obviously less data on the target domain for training.
Since the labeling work for activity recognition is not pixel-wise annotation and each sample only needs one label, the data collection may need more time than labeling one sample.
In that case, our training set on the target domain for the $20$-shot setting needs obviously less time compared with the UDA task on each DA setting.
On Sims4Action $\rightarrow$ TSH, $97.7\%$ of the samples from the target domain for the training are discarded in FSDA-AR compared with UDA, while FSDA-AR delivers a better performance.
The performance of FSDA-AR is comparable to UDA as mentioned before, which indicates that FSDA-AR is a more efficient setting, especially when the data collection in the target domain is hard to execute. 
We point out that FSDA-AR is an important research direction in the future and our work will serve as an important test bed in this direction.

\begin{table}[t]
\caption{Analysis of the sample number that is required for FSDA-AR and UDA under each setting.}
\label{tab:details}
\centering
\scalebox{1}{\begin{tabular}{l|rrrr|r} 
\toprule
\midrule
\textbf{ Setting} & \multicolumn{4}{c}{\textbf{FSDA}} & \textbf{UDA} \\
                  & \textbf{S-1}  & \textbf{S-5} & \textbf{S-10} & \textbf{S-20} &              \\
\midrule
UCF~\cite{soomro2012ucf101}$\rightarrow$HMDB~\cite{kuehne2011hmdb}          &      12         &      60        &        120       &      240         &       840       \\
HMDB~\cite{kuehne2011hmdb}$\rightarrow$UCF~\cite{soomro2012ucf101}          &        12       &        60      &        120       &       240        &       1438       \\
Sims4Action~\cite{roitberg2021let}$\rightarrow$TSH~\cite{das2019toyota}   &        10       &        50      &         100      &       200        &        8552      \\
D1$\rightarrow$D2~\cite{damen2018scaling}              &       8        &        40      &        80       &        160       &        2495      \\
D2$\rightarrow$D1~\cite{damen2018scaling}             &        8       &         40     &        80       &        160       &         1543     \\
D2$\rightarrow$D3~\cite{damen2018scaling}             &        8       &       40       &       80        &        160       &       3897       \\
D3$\rightarrow$D2~\cite{damen2018scaling}             &        8      &        40      &          80     &       160        &     2495         \\
D1$\rightarrow$D3~\cite{damen2018scaling}             &        8       &        40      &       80        &        160       &         3897     \\
D3$\rightarrow$D1 ~\cite{damen2018scaling}            &        8      &         40     &          80     &         160      &    1543\\          
\midrule
\bottomrule
\end{tabular}}

\end{table}

\end{document}